\documentclass[10pt,twocolumn,letterpaper]{article}

\usepackage[table]{xcolor}% http://ctan.org/pkg/xcolor
\usepackage{cvpr}
\usepackage{times}
\usepackage{epsfig}
\usepackage{graphicx}
\usepackage{amsmath}
\usepackage{amssymb}
\usepackage{pifont}% http://ctan.org/pkg/pifont
\newcommand{\cmark}{\ding{51}}%
\newcommand{\xmark}{\ding{55}}%
\usepackage[pagebackref=true,breaklinks=true,colorlinks,bookmarks=false]{hyperref}

\cvprfinalcopy % *** Uncomment this line for the final submission

\makeatletter
\@namedef{ver@everyshi.sty}{}
\makeatother
\usepackage{tikz}
\usetikzlibrary{arrows.meta}
\usetikzlibrary{positioning,matrix}
\newcommand{\etheta}{\widetilde{\theta}}

\begin{document}
%%%%%%%%% TITLE
\title{Does it work outside this benchmark? Introducing the Rigid Depth Constructor tool, depth validation dataset construction in rigid scenes for the masses}

%\titlerunning{Rigid Depth Constructor}        % if too long for running head

\author{\setlength{\tabcolsep}{25pt}
\begin{tabular}{lr}Clément Pinard & Antoine Manzanera \end{tabular} \\
        U2IS, ENSTA Paris, Institut Polytechnique de Paris\\
        828, Boulevard des Maréchaux, 91762 Palaiseau Cedex\\
        \begin{tabular}{lr}{\tt\small clement.pinard@ensta-paris.fr} & {\tt\small antoine.manzanera@ensta-paris.fr} \end{tabular}
        }


\maketitle
%%%%%%%%% ABSTRACT
\begin{abstract}
   We present a protocol to construct your own depth validation dataset for navigation. This protocol, called RDC for Rigid Depth Constructor, aims at being more accessible and cheaper than already existing techniques, requiring only a camera and a Lidar sensor to get started. We also develop a test suite to get insightful information from the evaluated algorithm. Finally, we take the example of UAV videos, on which we test two depth algorithms 
   %leading the benchmark for KITTI  
   that were initially tested on KITTI \cite{Uhrig2017THREEDV}
   and show that the drone context is 
   %vastly 
   dramatically
   different from in-car videos. This shows that a single 
   context
   benchmark should not be considered reliable, 
   %enough, 
   and when 
   %wanting to use a depth 
   developing a depth estimation
   algorithm, one should 
   %try to 
   benchmark it 
   %with a dataset that fits exactly his  
   on a dataset that best fits one's
   particular needs, 
   %even if it 
   which often
   means creating a brand new one. Along with this paper we provide the tool with an open source implementation and plan to make it as user-friendly as possible, to make depth dataset creation possible even for small teams.
   %AM A RELIRE - Recapitulatif des contributions du papier
   Our key contributions are the following: We propose a complete, open-source and almost fully automatic software application for creating validation datasets with densely annotated depth, adaptable to a wide variety of image, video and range data. It includes selection tools to adapt the dataset to specific validation needs, and conversion tools to other dataset formats. Using this application, we propose two new real datasets, outdoor and indoor, readily usable in UAV navigation context. Finally as examples, we show an evaluation of two depth prediction algorithms, using a collection of comprehensive (e.g. distribution based) metrics.
\end{abstract}

%%%%%%%%% BODY TEXT
\section{Introduction}

Using computer vision for navigation has long been well established, as a camera sensor is very easy to set up, cheap and power efficient. The main uses are for odometry and 
%depth 
3D
maps which are then used to control the navigation, 
%in order to avoid any potential obstacle.
especially find a path and avoid obstacles.

Estimating depth from a camera is not an easy task, and validation data is very hard to obtain. 
Indeed, knowing depth requires to know the 3-dimensional environment the camera is facing with respect to its position. This requires to 
%measure depth explicitely 
explicitly measure depth
with 
%active 
range
sensors like Lidar or ToF cameras.

A major example of depth validation dataset is KITTI \cite{geiger2012we}, where a set of cameras and a Lidar are mounted on
%the camera. Later,  
a car. Following the acquisition,
the Lidar and video signals are calibrated and synchronized in order to construct sparse depth maps for every camera at every moment. The main problem with this method is that you need to construct a rigid rig between a Lidar and a camera, which, in addition to 
%be 
being
very costly can become very heavy, and 
is
not suitable to recreate the natural movement of a handheld camera or a consumer UAV camera.

To 
%answer 
address
this problem, we propose a way to construct a depth dataset with a two-step method that first uses a Lidar to scan an environment and then localizes images from a video camera with respect to the Lidar point cloud. Our method aims at being the most user friendly possible, 
%with flexibility 
and with maximal flexibility, both
on the methods to construct the point cloud, and 
%flexibility 
on the type of camera used for acquisition.

Along with this paper, we provide the tool as an open source package thoroughly tested with an industrial research team to ensure its usability. We will present the general method that is widely inspired from ETH3D \cite{schoeps2017cvpr}, and present three use cases that we were able to construct. Finally, we will use these datasets to show with a benchmark on 
%depth from a single camera 
monocular depth estimation
algorithms that results can vary greatly depending on the context and that the in-car environment, well referenced 
%by KITTI 
through the KITTI benchmarks,
is far from being representative of all navigation use cases. This means that for 
%every 
each new
navigation scenario, a new dataset should be constructed, at least for validation, which is exactly what our tool 
%tries to make 
aims at making
easier and cheaper.

%-------------------------------------------------------------------------
\section{Related works}
\subsection{Depth estimation for navigation}
Depth estimation, and more generally 3D perception is a core task for autonomous navigation.
%, in order to avoid obstacles. 
In the context of very light vehicles such as UAV, that can't carry heavy hardware, depth is often deduced from one or multiple cameras. The stereo camera is often used to compute depth from disparity \cite{nalpantidis2009stereovision}, 
%but it can also be used 
that can also be estimated
with a single camera, which has the advantage of being much cheaper to integrate. Depth can then be deduced from motion, by using methods based on epipolar geometry \cite{hartley_zisserman_2004}. Structure from motion and SLAM techniques can then be used to deduce both depth and 
camera 
movement 
%only using 
using only
optical flow and geometric equations. It must be noted that all structure from motion algorithms 
%will 
require rigid scenes, 
i.e.
without moving objects.

%Simultaneously, 
More recently,
depth inference networks have been shown to be able to estimate depth from 
%only one 
a single
image solely from perspective and context. Indeed, with only one image, they were able to get reasonable scale invariant 
(i.e. relative depth maps)
quality measures \cite{FuCVPR18-DORN,lee2019big,saxena20083,eigen2014depth,zhou2017unsupervised}.
However, in a navigation context, the scale invariant quality is not really interesting without a way to link the estimation to the real world. 
%Any depth algorithm used for navigation needs to provide depth with absolute values instead 
Instead 
of relative or normalized
values, 
absolute depths are needed
for a navigation algorithm such as Model Predictive Control \cite{lopez2017aggressive} to be used.
%As such, 
To this end,
depth algorithms need to estimate both depth and movement in order to work for navigation. 
%In these examples, 
Among the examples mentioned above,
Zhou \etal \cite{zhou2017unsupervised} is the only one we can work with.
It is important to note 
that
depth from a single image does not use pixel displacements. It is deemed a more bio-inspired method which makes use of end-to-end training of convolutional neural networks and their capacity of generalizing implicit pixel structure, but it does not rely on 
%hard mathematics equations. 
elaborate geometric constraints.
As such, it is 
%to be 
expected 
%that they will 
to
be less robust for unusual scene, especially with a confusing perspective.

\subsection{Validation sets and benchmarks: the specific case of consumer UAV cameras, or why we need a new dataset}
This project was mostly motivated by the specific use case of depth estimation from UAV consumer drone, which lacks a proper validation dataset. Indeed, most depth algorithms are currently tested either for autonomous driving environments such as KITTI \cite{Uhrig2017THREEDV}, or indoor environment such as NYUv2 \cite{Silberman:ECCV12}. In these evaluation frameworks, single image depth prediction algorithms heavily prevail 
%in 
within
the leaderboard, which makes these techniques {\em de facto} state of the art. 

As shown 
%by Pinard \etal 
in
\cite{pinard2018learning}, the context of UAV navigation is much more heterogeneous than 
%both of 
these environments, both in terms of 
%orientation, movement and position 
camera pose
and surrounding environment. However, contrary to KITTI, the problem of moving objects is deemed (at least for now) a secondary problem when flying high above the ground. It is thus not certain that algorithms performing well on KITTI or NYUv2 will not perform poorly in this context. This is especially true when considering that moving objects in KITTI make depth from motion much harder, while depth from context is not robust enough to the viewpoint variability. As such, these datasets address issues for particular use cases with their own difficulties that are not reflected in the UAV use case and vice-versa. This idea is corroborated by the fact that the Sintel depth dataset \cite{Butler:ECCV:2012} remains largely unsolved for the moment because in addition to moving objects, scene heterogeneity is much more prevalent. 

Inspired by 
the
Flying Chairs 
synthetic dataset
for optical flow\cite{DFIB15}, which 
%focused more on having many samples than having realistic images,
is founded on a surrealistic abstraction of the difficulties of optical flow,
Still Box \cite{isprs-annals-IV-2-W3-67-2017} has been proposed. It is a synthetic dataset trying to recreate the difficulties of a UAV flight, focusing on heterogeneity 
of appearance,
%with 
using
random shapes and textures. As such, it is a good training dataset, the same way Flying chairs is a good training dataset for optical flow, but
obviously,
it is not suited for evaluation.
%In table \ref{tab:datasets} are compared each dataset difficulties and how no existing dataset that serves as a benchmark for depth quality 
Table \ref{tab:datasets}, which compares the difficulties of the datasets currently used as benchmarks for depth quality, shows that none of them
is relevant for the UAV camera use case.

\begin{table*}
    \resizebox{\textwidth}{!}{
    \begin{tabular}{c|c|c|c|c|c}
         & KITTI\cite{Uhrig2017THREEDV} & NYUv2\cite{Silberman:ECCV12} & Sintel Depth\cite{Butler:ECCV:2012} & Still Box \cite{isprs-annals-IV-2-W3-67-2017} & a good UAV dataset \\
         \hline
        Moving objects & \cmark & \xmark & \cmark & \xmark & \xmark \\
        Outdoor & \cmark & \xmark & \cmark & N/A & \cmark \\
        Camera orientation variation & \xmark & \cmark & \cmark & \cmark & \cmark \\
        Camera position variation & \xmark & \xmark & \cmark & \cmark & \cmark \\
        Real videos & \cmark & \cmark & \xmark & \xmark & \cmark
    \end{tabular}
    }
    \caption{Difficulties featured in 
    %each dataset
    existing datasets.
    Being non photo-realistic, the distinction between Indoor and Outdoor is irrelevant for Still Box.}
    \label{tab:datasets}
\end{table*}

\subsection{Constructing a depth enabled dataset}
The 
%main 
basic
principle of 
%dataset constructed with depth enabled 
depth enabled dataset construction
is to use a device 
with reliable depth estimation capability,
that won't be available during evaluation. 
%for which a more reliable depth estimation method exist. 
Typically, we can use a rig with a RGB camera and a depth sensor like structured light \cite{Silberman:ECCV12}, Time of Flight, embedded Lidar \cite{Uhrig2017THREEDV, diode_dataset}, or light-field camera grids \cite{Schilling_MG-BDDPBL}. For evaluation, only the camera will be available, and the evaluation step will then measure the agreement between "reliable depth" measured by the dedicated sensor and estimated depth. It is important to note that an evaluation is only informative up to a certain point, where the quality of both methods are comparable. For dedicated depth sensors, we usually rely on the vendor's datasheet. However, this solution requires that we can emulate the camera movement of the use case we are trying to cover, \ie without potentially heavy depth sensors. This is problematic for consumer UAV, because the typical movement of its camera is difficult to mimic. Not only the camera is well stabilized, with a very smooth trajectory, that is not reproducible by hand, but the size of these cameras make it easy to fly very close to obstacles, which is not reproducible by a heavier UAV that could carry an additional depth sensor. These kinds of UAVs are usually very dangerous and need to be operated far from obstacles. 

An interesting method for rigid scenes has been presented with ETH3D \cite{schoeps2017cvpr} and {\em Tanks and Temples} \cite{Knapitsch2017} that usually serve to evaluate photogrammetry. Instead of having a depth sensor and a camera attached to the same rig, the data acquisition is done in two steps. They first get a point cloud measure from fixed laser scanners such as the FARO Focus, and then take images from cameras in the same scene after removing the laser scanner. This implies that no object is moving in the scene, but allows for any camera to be used. In the particular case of ETH3D, they added a photogrammetry step, using the taken pictures and some RGBD rendering of the Lidar scanner with the COLMAP software \cite{schoenberger2016sfm}. This allowed them to get frame localization with respect to each other, but also with respect to the Lidar scanner position. With this technique, they were able to get depth map indirectly 
%with 
from
the Lidar point cloud. They then used the generated depth map and camera position to get a stereo validation, with rectified frames. The ETH3D dataset construction method has been used as a foundation
%for this article. 
of our framework.
Our goal is to generalize their method not only for photogrammetry oriented footage, but for all kinds of videos, using anything available to acquire a 
3D
point cloud of a scene.

\section{Dataset creation method}

\subsection{The foundations: COLMAP and ETH3D}
ETH3D already offers to compute depth maps from a Lidar point cloud and a 
%free 
separate
camera, but it has been used in a very particular context and some of its methods are not suited for a more general one. Because this article aims at using the same tools as ETH3D, we analyze them and indicate what aspects need to be changed in order to increase flexibility.

\subsubsection{Photogrammetry with COLMAP}
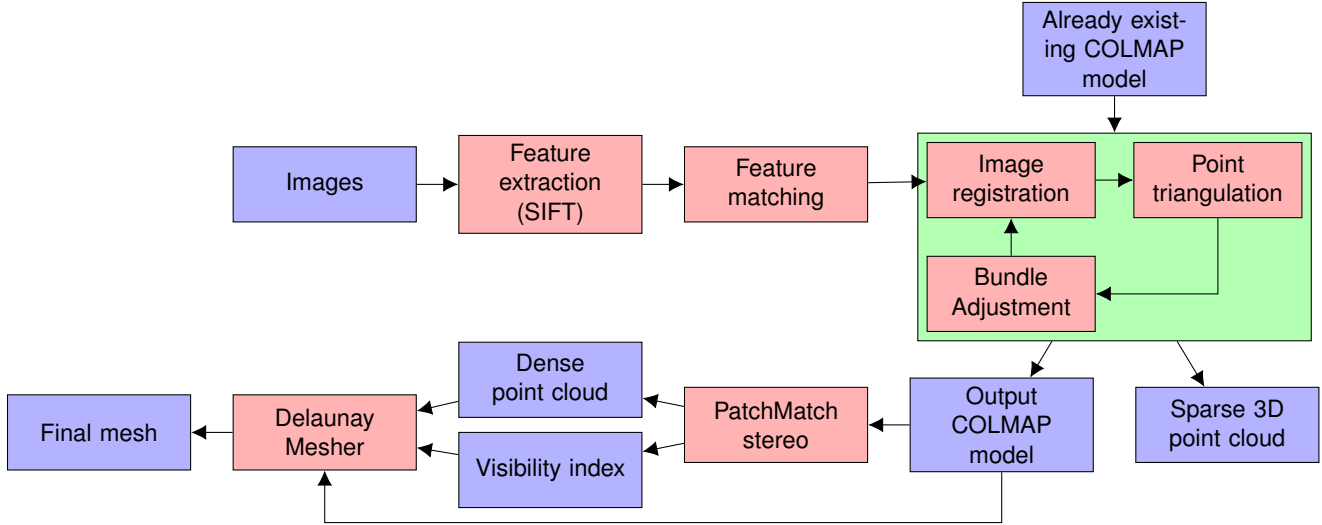
\begin{figure*}
\tikzset{%
  >={Latex[width=2mm,length=2mm]},
  % Specifications for style of nodes:
            base/.style = {rectangle,draw=black,
                           minimum width=0.1cm, minimum height=1cm, align=center},
            io/.style = {base, fill=blue!30},
            op/.style = {base, fill=red!30},
            mapper/.style = {base, fill=green!30},
            kmeans/.style = {base, fill=orange!15},
            network/.style = {base, fill=purple!30}
}
\begin{tikzpicture}[font=\sffamily\small]
\begin{scope}[node distance=3cm and 1cm]
  % Specification of nodes (position, etc.)
  \node (img)             [io,text width=2.2cm]              {Images};
  \node (fe)      [op, right of=img,text width=2.2cm]          {Feature \\ extraction (SIFT)};
  \node (fm)      [op, right of=fe,text width=2.2cm]   {Feature matching};
  \matrix (M) [matrix of nodes, mapper, right of=fm, xshift=1.5cm, yshift=-0.7cm, column sep=0.5cm, row sep=0.5cm, nodes={op, text width=2cm}]{
    Image registration & {Point \\ triangulation} \\
    Bundle Adjustment\\
  };
  \node (output)  [io, below of=M, text width=2.2cm, xshift=-1.5cm, yshift=0.5cm] {Output COLMAP model};
  \node (spm)      [io, below of=M,text width=2.2cm, xshift=1.5cm, yshift=0.5cm]   {Sparse 3D point cloud};
  \node (pms)      [op, left of=output,text width=2.2cm]   {PatchMatch stereo};
  \node (ecm)     [io, above of=M, text width=2.2cm, yshift=-0.5cm] {Already existing COLMAP model};
  \node (dpc)     [io, left of=pms, text width=2.2cm, yshift=0.6cm] {Dense point cloud};
  \node (vf)      [io, left of=pms, text width=2.2cm, yshift=-0.6cm] {Visibility index};
  \node (dm)      [op, left of=vf, text width=2.2cm, yshift=0.5cm] {Delaunay Mesher};
  \node (fmesh)      [io, left of=dm, text width=2.2cm] {Final mesh};
 
  \draw[->]     (img) -- (fe);
  \draw[->]     (fe) -- (fm);
  \draw[->]     (fm) -- (M-1-1);
  \draw[->]     (M-1-1) -- (M-1-2);
  \draw[->]     (M-1-2) |- (M-2-1);
  \draw[->]     (M-2-1) -- (M-1-1);
  \draw[->]     (M) -- (spm);
  \draw[->]     (output) -- (pms);
  \draw[->]     (ecm) -- (M);
  \draw[->]     (pms) -- (dpc);
  \draw[->]     (pms) -- (vf);
  \draw[->]     (dpc) -- (dm);
  \draw[->]     (vf) -- (dm);
  \draw[->]     (dm) -- (fmesh);
  \draw[->]     (M) -- (output);
  \draw[->]     (output) -- +(0,-1.3cm) -| (dm);
  \end{scope}
  \end{tikzpicture}
\linebreak
\caption{Photogrammetry workflow used in COLMAP. Note that the mapping process can stop at "Image registration" and the Delaunay mesher accepts any point cloud as long as the visibility index is correct.}
\label{fig:colmap}
\end{figure*}

COLMAP \cite{schoenberger2016mvs, schoenberger2016sfm}  is a photogrammetry tool designed to be very robust, in order to reconstruct a 3D model from 
%random 
crowdsourced
images taken 
%on 
from
the internet. To go from a set of images to a 3D model with images localization, the steps applied are presented in Fig.~\ref{fig:colmap}.
The main problem of COLMAP is that even if the matching process can be dramatically accelerated with vocabulary tree matching \cite{schoenberger2016vote}, 
%the mapping process is still 
it is still in
$\mathcal{O}(n^3)$ where $n$ is the number of images. As such, the reconstruction process can be very long for videos.

\subsubsection{Depth generation with ETH3D}
ETH3D \cite{schoeps2017cvpr} uses COLMAP to localize calibrated images with respect to a Lidar point cloud taken with a FARO Focus. The FARO Focus is a fixed point Lidar that renders colored 3D points. Since it is fixed, every 3D point is measured from the same origin. This device allows them to synthesize high quality 
%images with depth values with know position (the position of the scanner, \ie the cloud origin) 
depth-valued images with known position of the scanner (\ie the cloud origin)
that can be integrated in the COLMAP reconstruction process. The position of each image is thus known with respect to the point cloud. Each image then gets its position refined 
%with a system of 
by matching
feature points 
%matched from images and from point cloud image rendering 
between real images and rendered images from the colored point cloud
of an equivalent camera at the estimated position. However, the localisation part is not usable for our use case, because we want to be able to use
any
point cloud,
\ie
without colors. We can note that, as it is stated by their paper, the localization step can be done when simply constructing a 3D model with COLMAP and then register it with respect to the Lidar point cloud, with \eg ICP \cite{121791}.

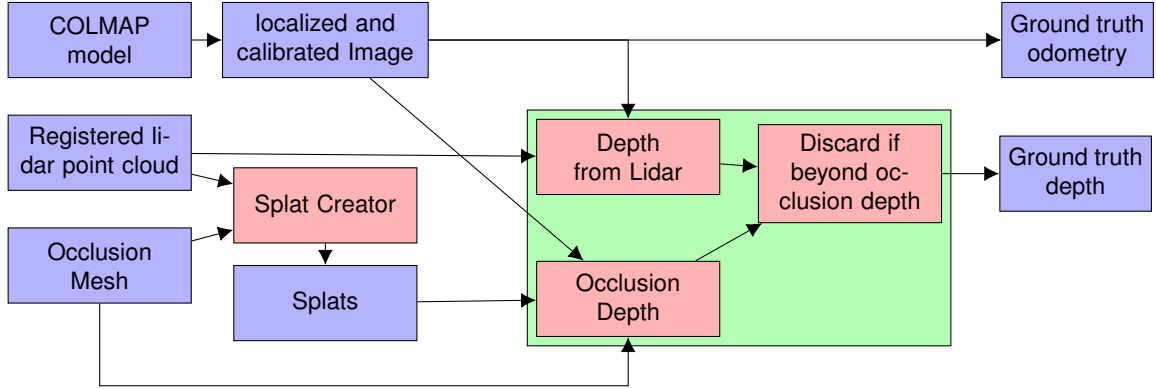
\begin{figure*}\centering
\tikzset{%
  >={Latex[width=2mm,length=2mm]},
  % Specifications for style of nodes:
            base/.style = {rectangle,draw=black,
                           minimum width=0.1cm, minimum height=1cm, align=center},
            io/.style = {base, fill=blue!30},
            op/.style = {base, fill=red!30},
            mapper/.style = {base, fill=green!30},
            kmeans/.style = {base, fill=orange!15},
            network/.style = {base, fill=purple!30}
}

\begin{tikzpicture}[font=\sffamily\small]
\begin{scope}[node distance=3cm and 1cm]
  % Specification of nodes (position, etc.)
  \node (cm)             [io,text width=2.2cm]              {COLMAP model};
  \node (img)   [io, right of=cm, text width=2.5cm] {localized and calibrated Image};
  \node (rlpc)  [io, below of=cm, text width=2.2cm, yshift=1.5cm] {Registered lidar point cloud};
  \node (oc)    [io, below of=rlpc,text width=2.2cm, yshift=1.5cm]          {Occlusion Mesh};
  \node (sc)    [op, right of=oc,text width=2.2cm, yshift=0.8cm]   {Splat Creator};
  \node (sp)    [io, below of=sc, text width=2.2cm, yshift=1.7cm] {Splats};
  \matrix (gtc) [matrix of nodes, mapper, right of=img, xshift=2.5cm, yshift=-2.5cm, column sep=0.5cm, row sep=0.5cm, nodes={op, text width=2.2cm}]{
    Depth from Lidar & Discard if 
    %below 
    beyond
    %AM: Check! (ya peut-être un truc qui m'échappe ?)
    occlusion depth \\
    Occlusion Depth\\
  };
  \node (gtd)     [io, right of=gtc-1-2] {Ground truth \\ depth};
  \node (odo)     [io, right of=img, xshift=7cm] {Ground truth \\ odometry};
  
  \draw[->]     (cm) -- (img);
  \draw[->]     (img) -| (gtc-1-1);
  \draw[->]     (img) -- (gtc-2-1);
  \draw[->]     (img) -- (odo);
  \draw[->]     (rlpc) -- (gtc-1-1);
  \draw[->]     (rlpc) -- (sc);
  \draw[->]     (sp) -- (gtc-2-1);
  \draw[->]     (oc) |- +(5,-1.6) -| (gtc-2-1);
  \draw[->]     (gtc-1-1) -- (gtc-1-2);
  \draw[->]     (gtc-2-1) -- (gtc-1-2);
  
  \draw[->]     (oc) -- (sc);
  \draw[->]     (sc) -- (sp);
  \draw[->]     (gtc-1-2) -- (gtd);
  \end{scope}
  \end{tikzpicture}
\linebreak
\caption{Photogrammetry workflow used with ETH3D. Note that this workflow does not include localization, which is much more complicated but is not available to us, 
%if we don't 
since we assume that we don't necessarily 
have a colored point cloud.}
\label{fig:eth3d}
\end{figure*}

Once the Images are localized with respect to Lidar, the depth rendering part is detailed in Figure~\ref{fig:eth3d}. Mainly, in addition to image calibration and localization, an occlusion mesh needs to be computed from the point cloud, which is then used to construct an occlusion depth. This depth map is 
%sensibly 
significantly
worse than the one from the point cloud, but it is only used to determine occlusion (and thus point visibility) and then avoid the ghosting effect of seeing through a 3D object 
%because 
due to the sparsity of
the point cloud. 
%is not dense enough. 
To this occlusion mesh we can add the "splats", which are created from isolated points: assuming 
%isolated
such
points are representative of thin objects (such as the leg of a chair), they may not be represented by the occlusion mesh but still count as occluding point, the splat creator is used to construct oriented squares at the point position, in order to avoid the risk of rendering 
%a 
depth values of a background object for the rest of the thin object.

In ETH3D\cite{schoeps2017cvpr}, the occlusion mesh is constructed with Poisson algorithm \cite{kazhdan2006poisson}. This supposes that point normals can be computed and oriented. Again, this is possible 
%from point cloud 
with point clouds
coming from a fixed laser scanner where normals are always oriented toward the origin, but it is not always the case for unstructured point clouds. We will 
then
have to find an alternative solution.

\subsection{Changing the ETH3D workflow}
%The 
The constraints from the ETH3D
use case context can be summarized this way:
\begin{itemize}
    \item The number of images is very 
    %low. If
    small. However if
    we want to use videos, we might have thousands of frames to localize.
    \item The Lidar 
    %scanner was 
    is
    a very high quality color-enabled fixed scanner. 
    %This means that A mesh can be computed much more easily than an 
    However we want to be able to use any collection of
    unordered point clouds,
    for which the computation of the mesh is much harder.
\end{itemize}

In short, we need to find a way to localize images with COLMAP in linear time while keeping a good 3D reconstruction, and we need a way to construct a good mesh from 
%an unordered point cloud.
unordered point clouds.

\subsubsection{Extending COLMAP reconstruction}

The issue regarding the number of images can be solved by simply using a subset of images for reconstruction. The reconstruction needs to be good enough to be precisely registered with respect to the Lidar point cloud. For 
%this 
a small
subset of frames, the full structure from motion mapping process can be applied even if it is very expensive. However, for remaining frames, the 3D point cloud will only be marginally better since 
%the view 
their views
are supposedly already covered by nearby frames. We can then afford to only register with respect to the already existing reconstruction. This process is much less expensive, as it does not need to 
%run 
perform
a global bundle adjustment 
%task 
at each 
%image.
frame.

The issue is now to choose a good subset of frames. This can be solved during the data acquisition step. Indeed, we can take a set of pictures dedicated to photogrammetry following guidelines in \cite{Photogrammetry}. The goal is, for a fixed number of pictures to be used in the mapping process, to have pictures with the most uniformly distributed position and orientation view points. This can be done for example with a UAV orbiting around a particular object or flying along a grid above the scene: pictures are sampled at a regular pace to ensure a good parallax between images.

\subsubsection{Constructing the Occlusion mesh}

In the hypothesis we were successful in localizing every image with respect to the Lidar point cloud, we can use a specific tool to construct a mesh. After point cloud densification with multi view synthesis and depth maps fusion, COLMAP outputs a point cloud with normals and a visibility index indicating from which frame each 3D point is visible. These two features are used for mesh reconstruction. Indeed, as discussed earlier, Poisson reconstruction \cite{kazhdan2006poisson} can be used thanks 
%for 
to
normals, and Delaunay meshing \cite{delauney} can be used with visibility index. If we make the unrealistic assumption that both dense reconstruction from COLMAP and Lidar point cloud are perfect, we can easily transfer these features from one cloud to another. Although COLMAP is known to have a low recall, 
by
discarding many points in textureless areas, in our experiments, we found that transferring feature from the nearest neighbor of each Lidar point was sufficient.

\subsection{Our final workflow}

\begin{figure*}\centering
\tikzset{%
  >={Latex[width=2mm,length=2mm]},
  % Specifications for style of nodes:
            base/.style = {rectangle,draw=black,
                           minimum width=0.1cm, minimum height=1cm, align=center},
            io/.style = {base, fill=blue!30},
            op/.style = {base, fill=red!30},
            mapper/.style = {base, fill=green!30},
            kmeans/.style = {base, fill=orange!15},
            network/.style = {base, fill=purple!30}
}

\begin{tikzpicture}[font=\sffamily\small]
\begin{scope}[node distance=3cm]
  % Specification of nodes (position, etc.)
  \node (lpc)             [io,text width=2.5cm]              {Lidar Point Cloud};
  \node (img)      [io, below of=lpc,text width=2.5cm, yshift=0.9cm]          {Photogrammetry Images};
  \node (fm)      [io, below of=img,text width=2.2cm, yshift=1.5cm]          {Frames metadata (if any)};
  \node (ftl)      [io, below of=fm,text width=2.2cm, yshift=1.5cm]          {Frames to localize};
  \node (os)     [op, right of=fm, text width=2.2cm]   {Optimal Sample};
  \node (sfm)      [op, right of=img,text width=2.2cm]   {COLMAP (SFM)};
  \node (geo)      [op, right of=sfm,text width=2.2cm]   {Geo registration};
  \node (dense)   [op, right of=geo,text width=2.2cm]   {COLMAP (Densification)};
  \node (dpc)     [io, right of=dense, text width=2.2cm, yshift=0.6cm] {Dense point cloud};
  \node (vf)      [io, right of=dense, text width=2.2cm, yshift=-0.6cm] {Visibility index};
  \node (icp)    [op, above of=dpc, text width=2.2cm, yshift=-1.5cm] {Registration (e.g. ICP)};
  \node (fr)     [op, below of=geo, text width=2.2cm, yshift=1.5cm] {Frame registration};
  \node (rpc)    [io, right of=icp, text width=2.2cm] {Registered Lidar point cloud};
  \node (vft)    [op, right of=dpc, text width=2.2cm, yshift=-0.6cm] {Visibility transfer};
  \node (dm)   [op, below of=vft, text width=2.2cm, yshift=1.5cm] {Delaunay mesher};
  \node (om)   [io, below of=dm, text width=2.2cm, yshift=1.5cm] {Occlusion Mesh};
  \node (lf)   [io, right of=fr, text width=2.2cm]{Localized Frames};
  
  \draw[->] (lpc) -- (icp);
  \draw[->] (img) -- (sfm);
  \draw[->] (ftl) -- (os);
  \draw[->] (fm) -- (os);
  \draw[->] (os) -- (sfm);
  \draw[->] (sfm) -- (geo);
  \draw[->] (geo) -- (fr);
  \draw[->] (geo) -- (dense);
  \draw[->] (dense) -- (dpc);
  \draw[->] (dense) -- (vf);
  \draw[->] (dpc) -- (icp);
  \draw[->] (icp) -- (rpc);
  \draw[->] (rpc) -- (vft);
  \draw[->] (dpc) -- (vft);
  \draw[->] (vf) -- (vft);
  \draw[->] (vft) -- (dm);
  \draw[->] (dm) -- (om);
  \draw[->] (ftl) -| (fr);
  \draw[->] (fr) -- (lf);
\end{scope}
\end{tikzpicture}
\linebreak
\caption{Simplified representation of our workflow before using ETH3D tools. Video frames get registered with respect to the reconstruction point cloud, along with the Lidar point cloud. As such, we can use ETH3D with the COLMAP model, the Occlusion mesh and the registered Lidar point cloud.}
\label{fig:workflow}
\end{figure*}
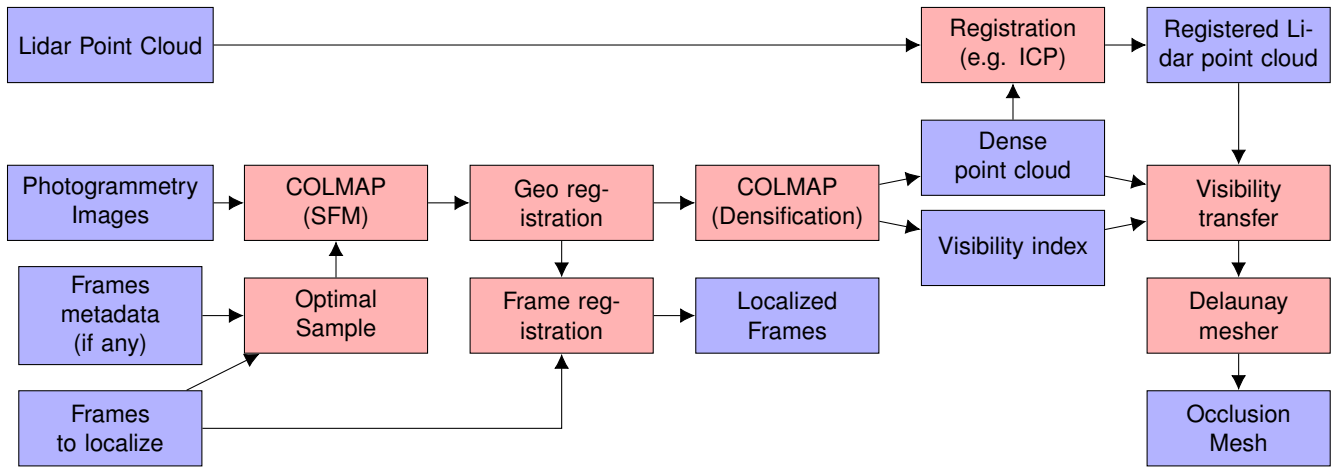

The final version of our workflow can be found in Figure~\ref{fig:workflow},
%. We then have an algorithm in order to have everything 
that represents the complete algorithm 
needed for ground truth creation from ETH3D. 
%which will be presented 
We now present it
in details.

\subsubsection{Data acquisition and ground truth point cloud creation}
In order for our protocol to be easy to follow, this part is very open and only asks for a point cloud of the scene,
%to measured by any means necessary, the point cloud is 
acquired by any mean, and
not necessarily colored. The point cloud will be used as a perfect ground truth, which means that subsequently constructed depth maps 
%will 
can
only be as precise as this point cloud. As suggested by \cite{schoeps2017cvpr}, a tripod fixed Lidar scanner can be used in order to have the maximum precision and density (precision at the millimeter level). However, other Lidar sensors can be used, such as mobile Lidars attached to a UAV, or a human handle. Although they lack 
%of 
precision (which is now at the centimeter level), their ease of use can be leveraged to have a much more complete point cloud, especially in cluttered environments or unreachable places like a building's roof. As a last resort, when no Lidar is available, even the result of a photogrammetry can be used, in order to compare a real-time depth algorithm to a thorough reconstruction algorithm that 
%makes no speed compromise on quality. 
does not sacrifice quality for speed.
Throughout this section, the point cloud that serves as a ground truth is referred to as Lidar point cloud, but everything applies the same for any other ground truth point cloud creation method.

\subsubsection{First thorough photogrammetry}
This first step uses COLMAP to construct a photogrammetry comprising a point cloud and the 
viewpoint
position of every image that was successfully localized. The reconstructed point cloud will then be localized with respect to the ground truth point cloud. Note that this step is different from ETH3D since we don't necessarily have a colored point cloud, and thus cannot synthesize images 
%with depth from lidar.
from Lidar data.

As said above, our goal is to make model reconstruction as efficient as possible. If we 
%were to add 
had to use
all the video frames we want to localize, the reconstruction would be comprehensive, but also extremely long. Instead, in addition to the "photogrammetry frames", we only include a subset of each wanted video to the photogrammetry process, so that the reconstruction is not too long 
%and enough 
while including sufficient number of
different view points. 
%are included. 
The sampling can be based on frame rate (e.g. only take one frame per second), but it is not ideal when the camera is not moving. In the case of frames with displacement metadata, we can sample a more efficient subset by using K-means \cite{Macqueen67somemethods} on a 6D point cloud composed of frames positions and orientation. Note that the importance of orientation in sampling can be parameterized by weighting the orientation coordinates in the 6D point cloud.

\subsubsection{Registration of ground truth point cloud with respect to the output of photogrammetry}

This step requires to find the optimal rigid transformation (rotation, translation and scale) between the reconstructed point cloud and the Lidar point cloud. Assuming COLMAP's reconstruction is good enough, a simple ICP \cite{121791} or 
%derivates 
related algorithm
like point-to-plane ICP \cite{CHEN1992145} can be used to align the two point clouds. It is probably the most 
%sensible 
sensitive
step of the process, and requires a human supervision.

Indeed, ICP is a somewhat unstable process that needs assessment and a good initialization. As such, a human needs to thoroughly check the point cloud alignment and manually estimate a first rough transformation, with \eg point pair picking. This can be done by \eg meshlab \footnote{\url{https://www.meshlab.net/}} or Cloudcompare \footnote{\url{http://www.cloudcompare.org/}}.

\subsubsection{Video localization}
This step only uses the image registration tool from COLMAP applied to the existing reconstruction. It does not contribute to the point cloud reconstruction, and only uses local bundle adjustment to localize the frame with its neighbors. As such, this operation is much faster than 
%if 
the whole reconstruction process, 
%with 
where
point triangulation and global bundle adjustment was used.

\subsubsection{Localization filtering}
The main interest of COLMAP compared to other SLAM method is that everything is global. This way every frame is not only localized with respect to its neighbors, but also with 
respect to any
other frame 
%that match enough key points with it. 
with which it shares a sufficient field of view.
As such, if each video frame contains features matched in enough well globally localized photogrammetry frames, the localization does not drift with respect to the reconstruction. However, since odometry is not perfect either, we can end up with a noisy trajectory that would not be possible 
%with 
for
a real camera 
%that has finite spatial and angular 
with bounded
acceleration. This is especially true for consumer UAVs which usually focus on 
%image 
video
smoothness 
%with 
using
a stabilized guimbal, for aesthetic purposes.

To reflect this observation, on each localized video, we apply a Savitzky-Golay filter \cite{doi:10.1021/ac60214a047} to both trajectory and orientation. This filter not only smoothes the movement, but also helps detecting outlier that were badly localized. This way, we can discard the frames for which the distance between estimated 
%6D position and filtered
and filtered 6D positions
is above a certain threshold, and interpolate their position from neighboring frames. These frames will not be used for depth evaluation, but can be used \eg for depth algorithms that rely on odometry.

\subsubsection{Depth and pose ground truth generation}
Finally, we use the ground truth creator developed for ETH3D \cite{schoeps2017cvpr} to construct depth maps for all the video frames. We now have a ground truth for odometry and depth for each successfully localized frame.

\subsubsection{Dataset conversion and evaluation subset creation}

Now that we have images with odometry and depth, we can convert the dataset 
%in more well known 
format in compliance with more popular
datasets. For example, we can use the same format as KITTI odometry, in order to ease 
%test 
validation
of depth algorithms on new datasets.

In addition, the same way Eigen\cite{eigen2014depth} proposed an evaluation split for depth, we can set a list of frames that will be used for depth evaluation. Odometry 
%lets us use the created dataset for 
can allow us to adapt our subset and its format to
several evaluation scenarios:
\begin{itemize}
    \item We can filter candidate frames 
    %by 
    according to the
    movement, \eg only forward motion (like in the context of a car), or without rotation.
    \item For algorithms based on normalized 
    (relative)
    depth with corresponding pose estimation with respect to previous frames (\eg SFMLearner \cite{zhou2017unsupervised}), we can solve the scale factor with displacement magnitude, as suggested in \cite{pinard2018learning}.
    \item For algorithms that need odometry, such as multi view stereo, or algorithms that need frames with compensated 
    %orientation 
    rotation
    \cite{pinard2018learning}, it can be provided. This scenario is realistic for navigation context in the case of a UAV, where velocity and orientation need to be known primarily for a stable flight and a smooth video, and thus are available for these algorithms.
\end{itemize}

\subsection{Automation}
All the above processes have been included 
%in 
into
a script that makes extensive use of COLMAP, ETH3D, PCL, and Parrot's Anafi SDK. This script is intended to be as easy to use and as flexible as possible, in order to cover 
%as many 
a wide range of 
use cases and budgets. 
%as possible. 
It is open-sourced on Github with extensive usage documentation. \footnote{\url{https://github.com/ClementPinard/depth-dataset-builder/}}

\subsection{How good 
%of 
a ground truth can we hope to reach?}

During the whole process, we have assumed that the 3D point cloud was perfect, or at least not improvable. We chose to trust precision ranges given by scanner vendors. For example for a fixed Lidar scanner such as the Faro Focus, the precision is below the centimeter, while for handheld Lidars such as Velodyne VLP16 used either with a UAV or handheld, the precision is below 5 centimeters. This means that depending on the device used for mapping, one needs to pay attention to the depth ranges seen during videos to 
%localize. 
be localized.
This is particularly true for videos very close to obstacles, where the precision of depth maps measured by COLMAP's patch match stereo step can be better than Lidar reference.

Using fixed or mobile scanners is then a trade-off between mobility, 
%how fast a particular area can be mapped, 
scan time,
precision and completeness.

\begin{figure*}
    \centering
    %\begin{tabular}{c}
     %    \includegraphics[width=0.45\textwidth]{images/colmap_problem.png}\\
      %   \includegraphics[width=0.45\textwidth]{images/colmap_no_problem.png}
    %\end{tabular}
    \begin{tabular}{cc}
         \includegraphics[width=0.49\textwidth]{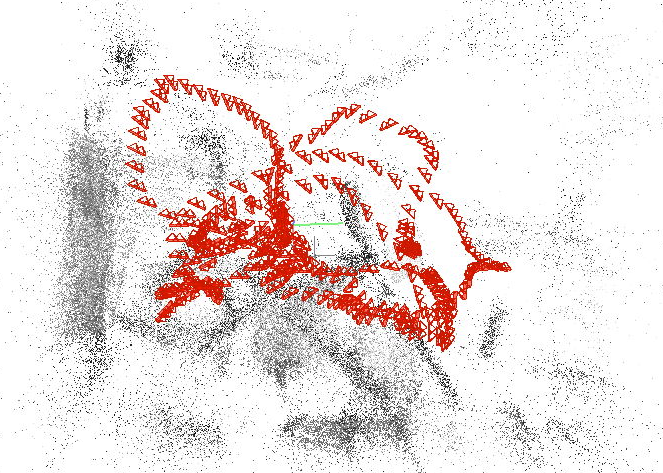} &
         \includegraphics[width=0.49\textwidth]{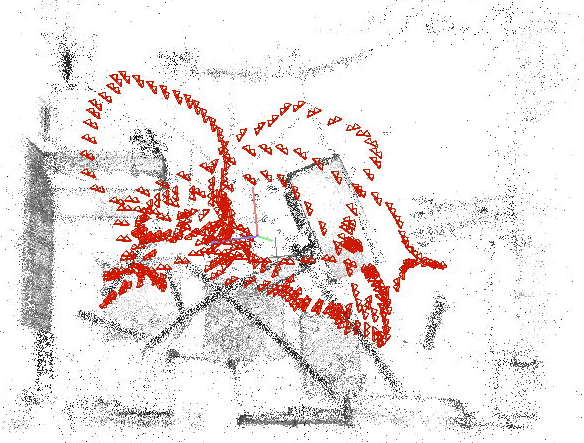}\\
         (a) & (b)\\
    \end{tabular}
    \caption{
    %subjective result 
    Visual qualitative assessment
    from COLMAP mapping process. 
    %Top: 
    (a)~localization 
    %was obtained 
    from available ground truth odometry, measured from Lidar and IMU. 
    %Bottom: 
    (b)~localization 
    %was 
    deduced by COLMAP during the mapping process 
    %via 
    with
    SLAM. Visualization was done via COLMAP GUI.}
    \label{fig:euroc_prob}
\end{figure*}

\begin{figure}
    \centering
    \def\arraystretch{0.5}
    \setlength{\tabcolsep}{1pt}
     \begin{tabular}{cc}
     \includegraphics[width=.59\columnwidth]{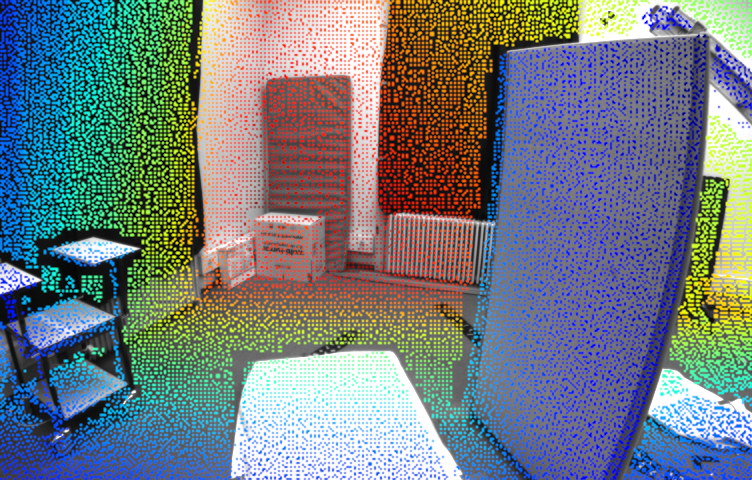} & 
      \includegraphics[width=.39\columnwidth]{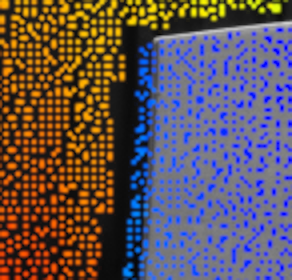}\\ (a) & (b)
      \end{tabular}
     \begin{tabular}{c}
     \includegraphics[width=\columnwidth]{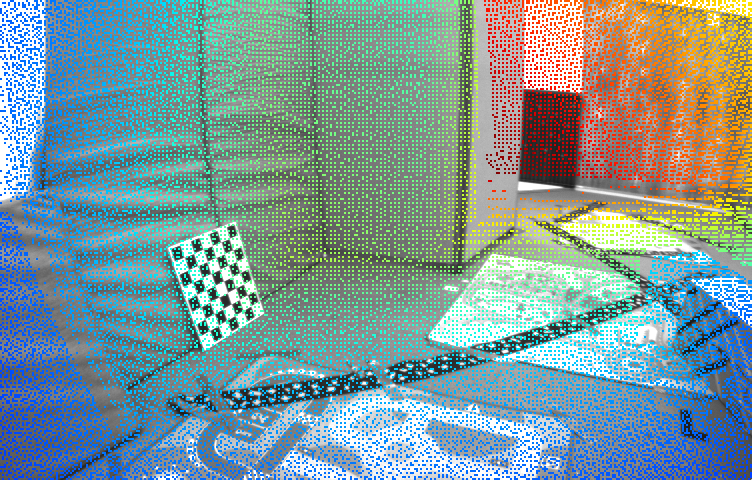}\\
     (c)
     \end{tabular}
    \caption{Depth map proposed by \cite{Gordon_2019_ICCV}. Image and colored depth maps are superposed to show the alignment problem, especially for foreground and background delimitation. %Top: 
    (a)~depth 
    map shown in Figure SM2 of \cite{Gordon_2019_ICCV} (from their supplementary materials), with 
    (b)~zoom 
    details.
    %Bottom: 
    (c)~depth 
    map from another sequence, where alignment error is larger than 40 pixels.}
    \label{fig:wild}
\end{figure}

Regarding odometry, we consider the localization made by COLMAP to be perfect. It can be noted that this was not the case for ETH3D, where they applied a pose fine tuning for each frame. Unfortunately, their method is only available for colored point clouds, which is not the case for most mobile Lidars. This problem can be mitigated by two factors:
\begin{itemize}
    \item The colorless scanners are also the 
    %ones that are less precise. 
    less precise ones.
    As such, solving this problem might get negligible improvement since the point cloud quality will then 
    %get the limiting factor for quality. 
    become the limiting factor.
    Otherwise, we can simply apply 
    %their 
    ETH3D's
    pose refinement technique.
    \item We tested COLMAP on EuroC dataset, and were surprised to see that odometry from 
    %given groundtruth 
    the provided ground truth
    (measured with a Lidar scanner and an IMU) was not very good compared to odometry computed by COLMAP. This can be seen on Figure \ref{fig:euroc_prob}, where the triangulated 3d points are visibly much less noisy from COLMAP odometry than from ground truth. This is corroborated with the EuroC depth dataset proposed by \cite{Gordon_2019_ICCV}, where synthesized depth maps from frame position and camera calibration were not exactly aligned with the camera, even in their illustrating figure (see Figure \ref{fig:wild}).
\end{itemize}
%As such, we have a good subjective reason 
In sum, we have good, albeit subjective, reasons
to think that the odometry from dedicated sensors is not necessarily needed compared to the one computed by COLMAP. We can only make a subjective manual assessment, but as mentioned in the first point, we believe that point cloud quality is often the limiting factor. However, we cannot ignore the fact that for some particular cases, such as a video that is isolated in a cluttered part of a scene and only connected to the rest of the photogrammetry by a few frames, the odometry can drift. This makes these localized frames misaligned with the Lidar point cloud, and thus 
%have 
with 
a poor ground truth depth, even if the Lidar point cloud registration step is optimal, because it's only a rigid transformation. A solution to this problem could be to apply a non-rigid registration \cite{6361384} of the COLMAP point cloud, deforming the cloud and 
%thus 
then
also the frame localization to fit the Lidar point cloud more precisely. 
%than with a simple rigid transformation. 
This might be the occasion of a future work combining COLMAP's bundle adjustment and cloud-to-cloud distance between COLMAP and Lidar point cloud. However, a more direct way of limiting this problem is to ensure during data acquisition that all video frames can be localized with a large number of photogrammetry oriented pictures, so that a "loop closure" step is applied very often, with pictures designed to have a very precise localization with respect to the reconstruction cloud.

\section{Measuring depth quality for navigation purposes}
Since we 
%are interested in constructing 
want to build
an evaluation dataset, we are 
%also interested in 
looking for
the most informative metrics. As we are mostly interested in the context of navigation, we chose to follow the methodology proposed
%by Pinard \etal 
in
\cite{pinard2018learning}. Namely, contrary to the well used "Eigen-split" \cite{eigen2014depth}, using metrics from Garg \etal \cite{garg2016unsupervised}, we want to apply a realistic navigation scenario, 
%and thus require 
that requires
the estimated depth 
%to evaluate 
to be absolute, and not up to a factor computed with the ratio of medians. This makes depth from single view algorithms unable to compete, unless there is a way to compute the scale factor by comparing odometry with actual displacement, which is much more realistic for any kind of autonomous vehicle. Single frames algorithms such as Deep Ordinal Regression Network (DORN)~\cite{FuCVPR18-DORN} or Big to Small (bts)~\cite{lee2019big} can thus not be used, but some other algorithms such as SFMLearner and its 
%derivates~
variants~
\cite{zhou2017unsupervised, monodepth2} are trained with a pose estimator and thus can be evaluated.

\subsection{On the information a metric gives}
Most of the time, for evaluation datasets, metrics are used for the only purpose of ranking different algorithms. This is useful for benchmarking and choosing an existing algorithm for one's usage, but it raises some problems from the end user's perspective:
\begin{itemize}
    \item It does not reflect the context of the use case that might be different from the original validation set. For example, if someone wants to estimate depth from a single camera but only for close objects because the long range is already covered by another sensor, this user won't be interested in long range depth estimation quality, and the metric used for benchmarking will have to reflect this.
    \item Some characteristics of an algorithm are inherently 
    %adversarial, 
    antagonistic,
    and thus a trade-off must be decided between 
    %the two. 
    them.
    The most usual example is accuracy {\em vs} speed. By ranking algorithms with only one metric, the dataset makes the trade-off decision in place of the end user and thus takes the risk of not being informative. For example, because speed is only given as declared from the authors, KITTI depth benchmark \cite{Uhrig2017THREEDV} completely disregards it. It would be interesting to have data presented into a 2D chart, the same way as VOT \cite{VOT_TPAMI} so that the end user can eventually 
    %decide the 
    set his own
    speed/accuracy trade-off. 
    %by himself.
    \item A metric can give information about the distribution of expected values given a particular estimation. As such, as shown 
    %by Pinard 
    in~
    \cite[p.~38]{pinardphd} all metrics are not the same when it comes to trying to characterize possible real values of an estimation. The ideal solution would be to give for each estimated depth value the exact distribution of real depth. That would require an infinite set, but the validation set can give an approximation that gives more insight than just a single number.

\end{itemize}

To reflect those 
%usages, 
practices,
we decide to have both a set of classic metrics and histograms to give as much information as possible for a given algorithm. Again, the code for metric measurement is open sourced on Github \footnote{\url{https://github.com/ClementPinard/depth-dataset-builder/tree/master/evaluation_toolkit}}. It consists in getting all depth pixels and their estimation, within an unordered set $V$. Note that this evaluation set is not image-wise, all depth values are collected at first with corresponding metadata, and the metric computation is done at the end. This is useful for images with a very sparse depth, where no representative statistics can be computed.

\subsection{Scalar metrics}
A scalar metric is obtained by computing a depth 
%error for each point and the meaning 
errors for all points and then averaging
them. As mentioned above, the mean is global
over the validation set $V$:
\begin{equation*}
\mathcal{E}_f = \frac{1}{|V|}\sum_{(\etheta, \theta)\in V} f(\etheta, \theta) 
%\simeq	 \mathbb{E}(f(\etheta, \theta ))
= \mathbb{E}_V(f(\etheta, \theta ))
\end{equation*}

Where $\theta$ is the ground truth depth for a particular pixel, 
%and 
$\etheta$ its estimate, 
and $f$ the error function.

Table \ref{tab:error_functions} shows provided error functions and corresponding names.

\begin{table*}
    \renewcommand{\arraystretch}{1.5}
    \centering
    \begin{tabular}{|l|c|c|}
    \hline
    Error Name & Acronym & 
    %Equation $f(\etheta, \theta)$
    $\mathbb{E}(f(\etheta, \theta ))$ \\ \hline
     Mean Absolute Error & $MAE$ & $\mathbb{E}|\etheta - \theta |$ \\ \hline
     Mean Relative Error & $MRE$ & $\mathbb{E}\frac{|\etheta - \theta|}{\theta}$ \\ \hline
     Mean Log Error & $MLE$ & $ \mathbb{E}|\log(\etheta) - \log(\theta)|$
     \\ \hline
     Standard Absolute Error & $SAE$ & $\sqrt{\mathbb{E}(\etheta - \theta )^2}$ \\ \hline
     Standard Log Error & $SLE$ & $\sqrt{\mathbb{E}(\log(\etheta) - \log(\theta))^2}$ \\ \hline
     Precisions $\delta$ & $P_\delta$ & $P\left(\left|\log\left(\frac{\etheta}{\theta}\right)\right| \leq \log(\delta) \right)$ \\ \hline
    \end{tabular}
    \caption{Considered 
    %Losses 
    Metrics
    Summary}
    \label{tab:error_functions}
\end{table*}

\subsection{Histogram metrics}

Histogram metrics cannot be used for ranking algorithms but they give much more information. We propose two different histograms:
\begin{itemize} 
    \item Depth wise error: Given a scalar metric, we can compute the error for particular depth values. 
    %$H(\theta_0) = \mathbb{E}_{\theta = \theta_0}(f(\theta, \etheta))$
    $$
    %H_(\theta_0) 
    H_D(\theta_0)
    = \mathbb{E}_{\{ \theta = \theta_0 \}}(f(\theta, \etheta))$$
    
    \item Difference distribution: This 
    %is a 
    distribution 
    %that 
    is normally centered around $0$, and its standard deviation is the standard error mentioned 
    %above. 
    on Table~\ref{tab:error_functions}.
    Having the whole distribution is especially interesting for distribution that are not symmetrical. A safety interval can then be deduced from this histogram for each side of estimated depth. For this part, we follow the assumptions that the log of depth estimation is more likely to follow a symmetrical distribution than raw depth estimation. It is equivalent to get a distribution of the ratio between 
    %ratio and groundtruth, 
    estimation and ground truth,
    usually centered around 1.
    
    $$ 
    %H(\alpha) =  P(log(\etheta) - log(\theta) = \alpha) = P\left(\frac{\etheta}{\theta} = e^\alpha\right) $$
    H_{\Delta}(\delta) =  P(\log(\etheta) - \log(\theta) = \delta) = P\left(\frac{\etheta}{\theta} = 10^\delta\right) $$
    
\end{itemize}

\subsection{Displacement wise metrics}
In addition to those metrics, we propose to have a displacement wise metric in the form of an histogram. Namely, if we know the displacement, we can deduce what point in the image the camera is moving toward to. 
%This is called the flight path vector FPV). 
This point, called the flight path vector (FPV) for aircraft, also corresponds to the epipole for multi-view geometry, and to the focus of expansion (FOE) for optical flow in the case of rotation-less movement. As a consequence, these points are deemed more important than the other ones.

$$\mathcal{E}_{FPV}(\alpha) = \mathbb{E}_{V(\alpha)}(f(\etheta, \theta ))$$

Where $V(\alpha)$ is the set of points at a distance $\alpha$ from the FPV of each image (be it in pixels, or in radians).

This particular distribution can help discard an algorithm that fails to estimate 
depth around
those points while having good metrics otherwise. This is the case for optical flow based method for a stabilized camera without rotation: optical flow is too 
%low 
small
around the FPV and thus 
%optical flow 
disparity based depth estimation
becomes too noisy. \cite[p~32]{pinardphd}

\section{Applications}

\subsection{Preamble}

In this section, we present two use cases we have covered using this tool. They both feature drone footage, and 
%we tested them with 
are used to test 
two algorithms: DepthNet \cite{isprs-annals-IV-2-W3-67-2017} and SFMLearner \cite{zhou2017unsupervised}. 
%Although they 
They
are not state of the art on 
%regular 
popular
benchmarks like KITTI\cite{Uhrig2017THREEDV}, 
%we have chosen these two algorithms for multiple reasons:
but they are relevant to demonstrate the interest of the proposed evaluation framework and data sets, for the following reasons:
\begin{itemize}
    \item They are both compatible with our evaluation scenario focused on navigation, because we can scale their depth estimation only using odometry
    \item They are supposed to be real time, contrary to MVS algorithms like the one used in COLMAP
    \item %They show similar metrics 
    Their performance metrics are similar
    on KITTI depth \cite{Uhrig2017THREEDV}
    \item Contrary to SFMLearner, DepthNet has been specifically designed to be robust
    to variability of context and scene layout. 
    As such, for heterogeneous data sets, we should expect DepthNet to perform better than SFMLearner
    \cite{schoenberger2016mvs}.
\end{itemize}

%For both evaluation scenarios, we share the same protocol. Since DepthNet and SFMLearner both relying on training a neural network, we 
The evaluation protocol is the same for the two scenarios. We
divide the data set into two parts, one for training and one for test. Images and depth maps are rescaled to $416 \times 234$, and for both neural networks, we follow their publication training schedule. DepthNet is pre-trained on the Still Box dataset \cite{isprs-annals-IV-2-W3-67-2017}, 
%and is trained with depth supervision, while 
then trained on the learning data set with photometric depth auto-supervision, 
%with 
and
supervision for odometry rotation.
SFMLearner is not pre-trained (we did not see improvements when pre-training with KITTI); %and
it is 
entirely
trained 
on the learning data set
without any supervision.

\subsection{First application: The drone Manoir dataset}
\subsubsection{Context}
\begin{figure}
     \centering
     \def\arraystretch{0.5}
    \setlength{\tabcolsep}{1pt}
     \begin{tabular}{cc}
          \includegraphics[width=.49\columnwidth]{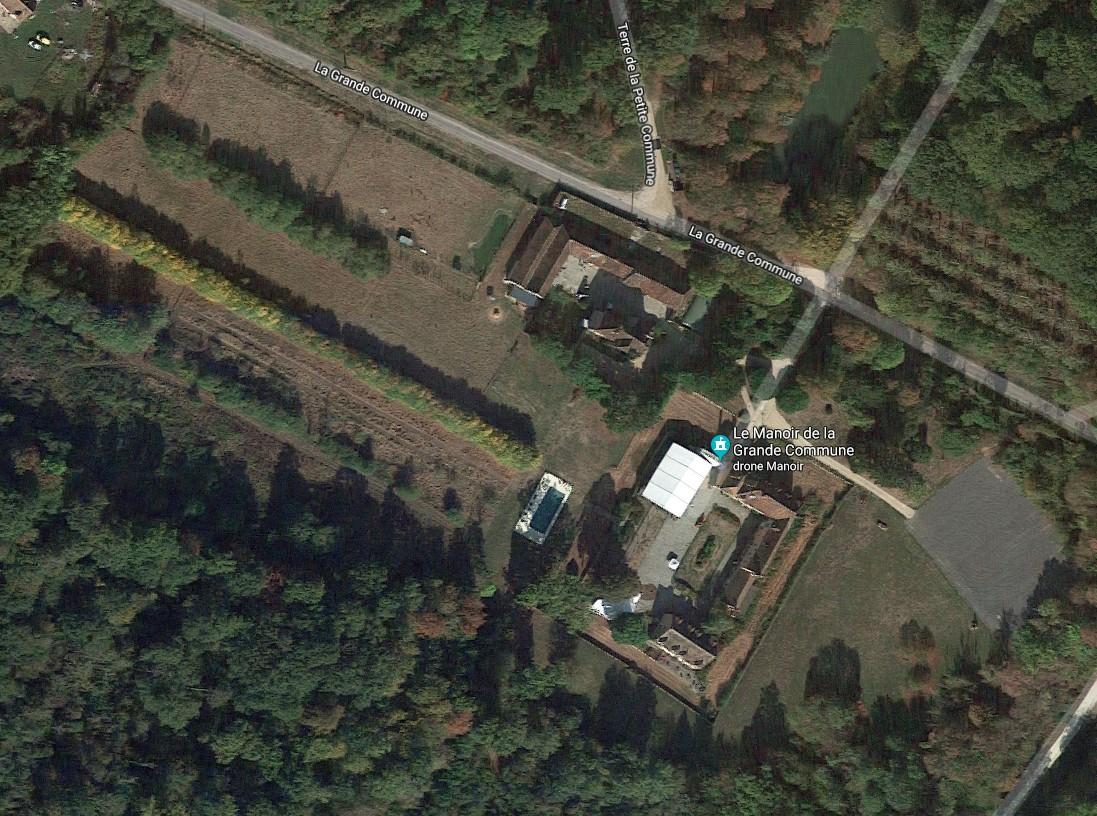} & \includegraphics[width=.49\columnwidth]{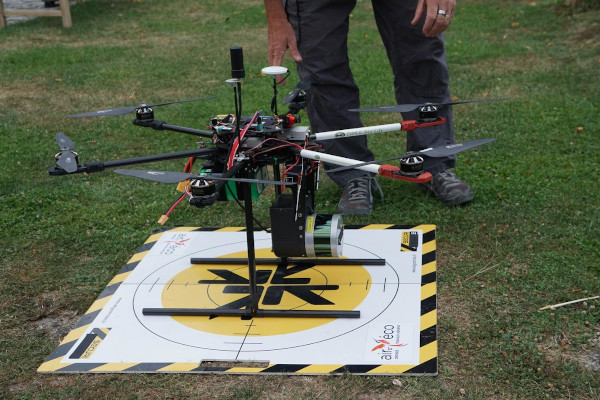} \\
          \includegraphics[width=.49\columnwidth]{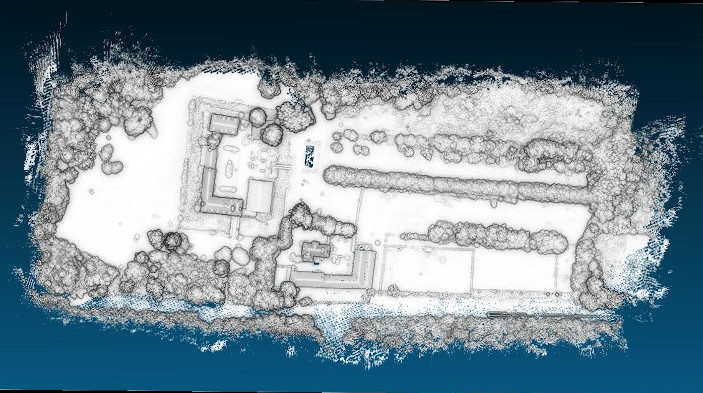} &
          \includegraphics[width=.49\columnwidth]{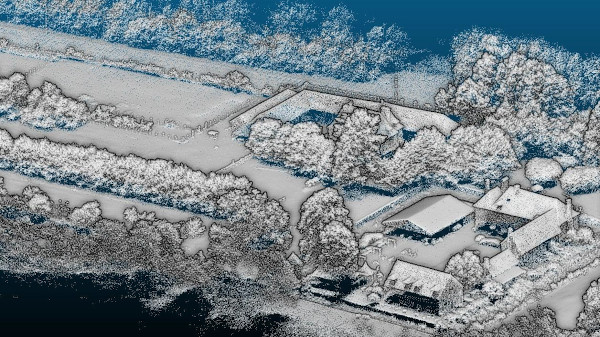}
     \end{tabular}
     \caption{3D Mapping process of the Manoir dataset. %From left to right, top to bottom: 
     Top:
     map of the 
     %area mapped, 
     scanned area
     and
     scan device used for mapping.
     %, and 
     Bottom: two views of the
     resulting point cloud.}
     \label{fig:manoir_map}
 \end{figure}

Our first 
%attempt at using this construction technique 
use case
is a scene with a mansion 
%in french countryside (manoir in french) with a terrain of size 350m by 100m. 
%(manoir in French) 
(in French: manoir)
in the countryside, on a $350 \times 100 m^2$ terrain.
The maximum altitude of obstacle is $20 m$. 3D Lidar data was captured by a DJI Matrice 600 with a Velodyne VLP-16 on board, with RTK GPS system (see figure~\ref{fig:manoir_map}). The flight altitude of this UAV was 30 meters at minimum for safety reasons. The UAV was used because it can cover an area much faster than any fixed scanner, and can easily scan building roofs. The whole scanning process took less than 10 minutes, meaning we could have covered a very large area in less than a day.

\begin{figure}
     \centering
     \def\arraystretch{0.5}
    \setlength{\tabcolsep}{1pt}
     \begin{tabular}{cc}
          \includegraphics[width=.49\columnwidth]{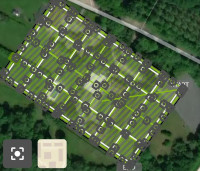} & \includegraphics[width=.45\columnwidth]{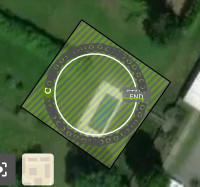} \\
          \includegraphics[width=.49\columnwidth]{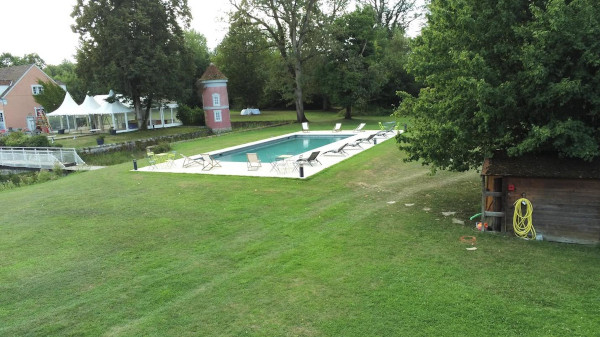} &
          \includegraphics[width=.49\columnwidth]{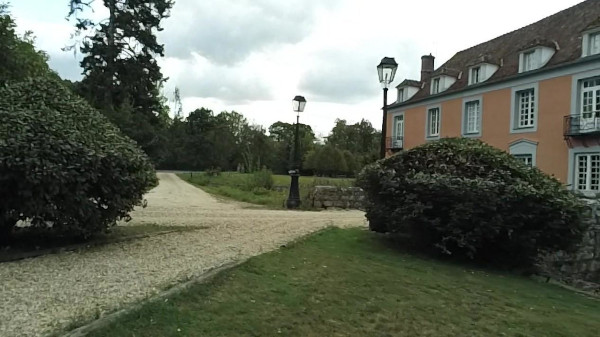}
     \end{tabular}
     \caption{Video acquisition process of the Manoir dataset. Top: grid and orbit flight plan used for photogrammetry. Bottom: Video samples from the Anafi drone, 4k (left) and 720p (right)}
     \label{fig:manoir_photo}
 \end{figure}

For photogrammetry oriented pictures, we used an Anafi drone with the free Pix4D app that lets us make one grid and two orbits above the field we wanted to scan. We also used a personal DSLR (Sony alpha-6000) for additional photo. We took videos at two different quality settings for a total of 65k frames to localize: 
 \begin{itemize}
     %\item 3840x20160 (4K) at 30fps, best quality setting.
     %\item 1280x720 at 120 fps, bad quality but high framerate.
     \item $3840 \times 2160$ (4K) at 30 fps, best quality setting.
     \item $1280 \times 720$ at 120 fps, bad quality but high frame rate.
 \end{itemize}
 See Figure~\ref{fig:manoir_photo}.
 
\subsubsection{Result}
\begin{figure}
    \centering
    \includegraphics[width=.49\columnwidth]{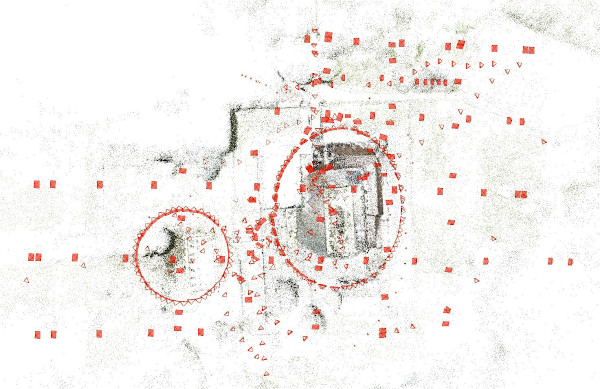}
    \includegraphics[width=.49\columnwidth]{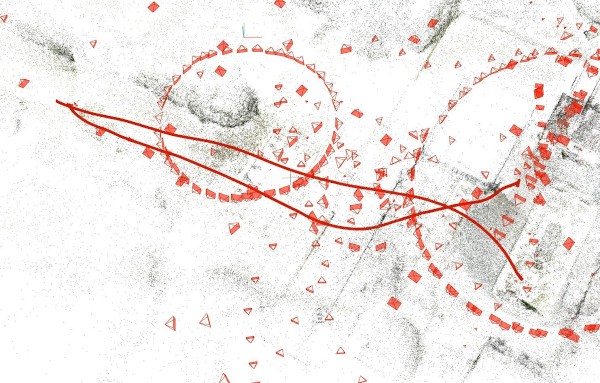}
    \caption{Left: global photogrammetry result 
    (point cloud in black, camera odometry in red). 
    Right: 
    %localization of all the frames of a given video in 
    camera odometry during a given video footage (red curve), w.r.t.
    the global photogrammetry.}
    \label{fig:manoir_photo_result}
\end{figure}

\begin{figure}
    \centering
    \includegraphics[width=\columnwidth]{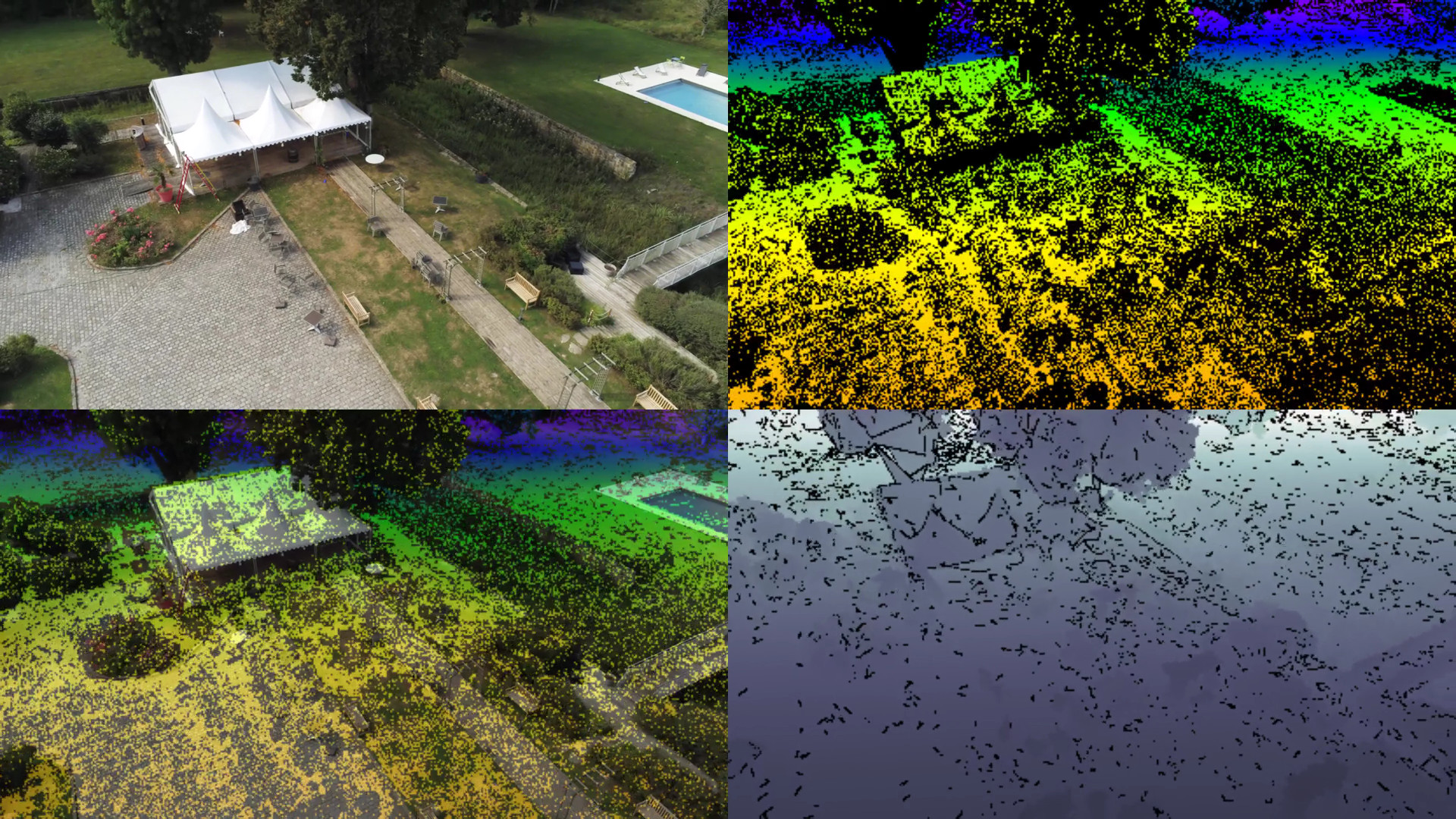}
    \caption{Depth result from the Manoir dataset. From left to right, top to bottom: RGB image, depth map, RGB + depth, occlusion depth.}
    \label{fig:manoir_full}
\end{figure}

%You can see a 
A
subjective result of photogrammetry
can be seen on
Figure~\ref{fig:manoir_photo_result}, for the optimal subset of 1000 frames and a full video taken by a drone. Finally, Figure~\ref{fig:manoir_full} shows a sample of computed depth maps. A video playlist is available as a supplementary material for all the sequences \footnote{\url{https://youtube.com/playlist?list=PLMeM2q87QjqhYA_LfJY925ZAGyD5cS6Q-}}

\subsubsection{Note on point cloud completeness}
\begin{figure}
    \centering
    \def\arraystretch{0.5}
    \setlength{\tabcolsep}{1pt}
    \begin{tabular}{cc}
        \includegraphics[width=.33\columnwidth]{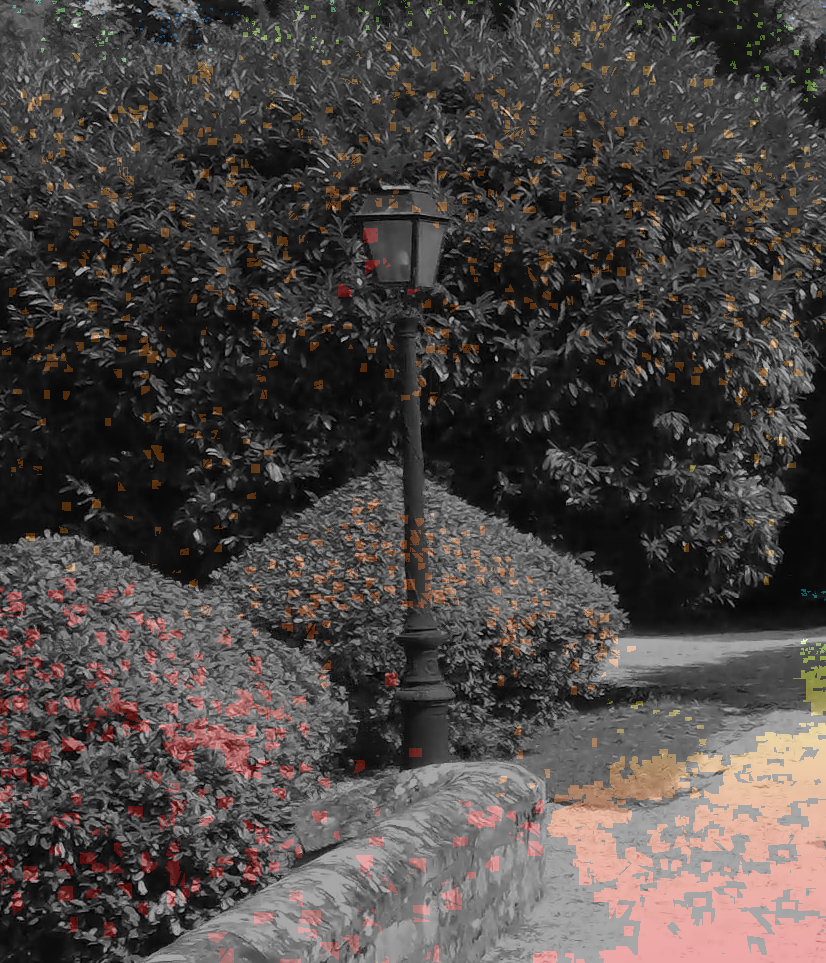} & \includegraphics[width=.66\columnwidth]{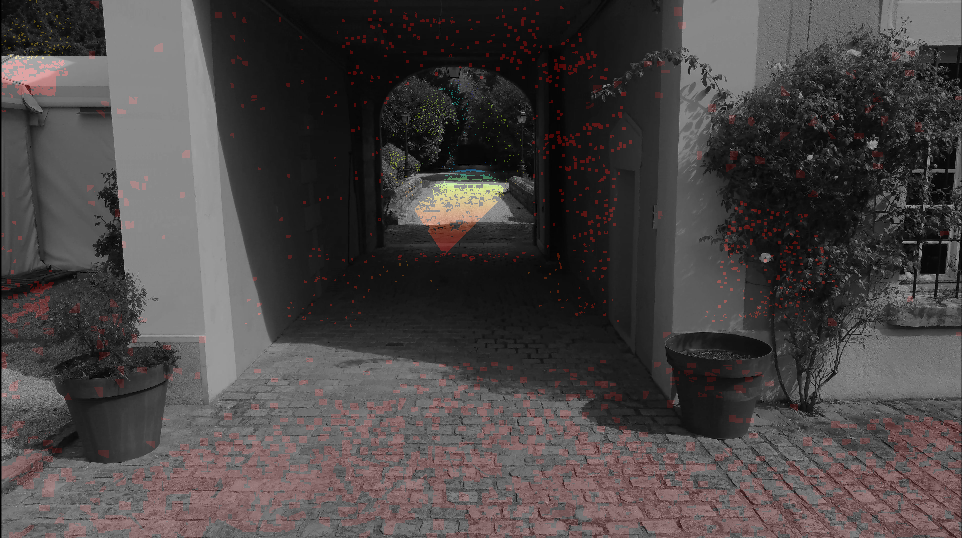} \\
        \includegraphics[width=.33\columnwidth]{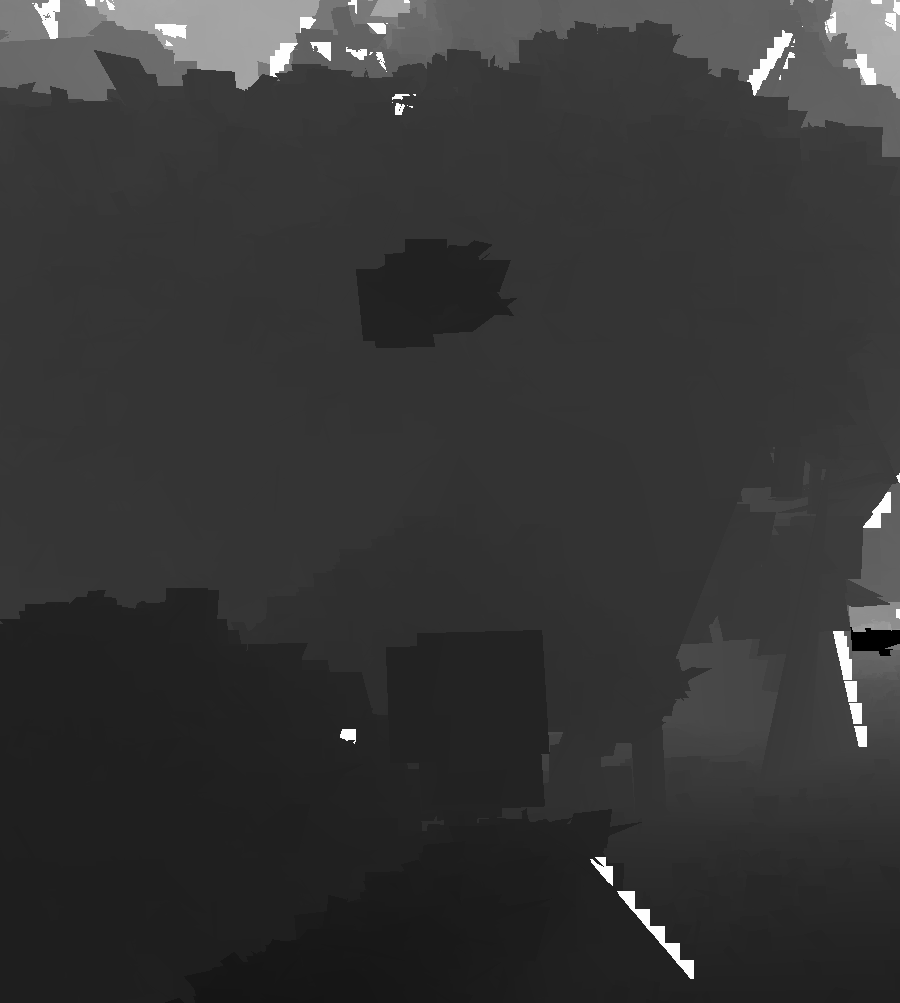} & 
        \includegraphics[width=.66\columnwidth]{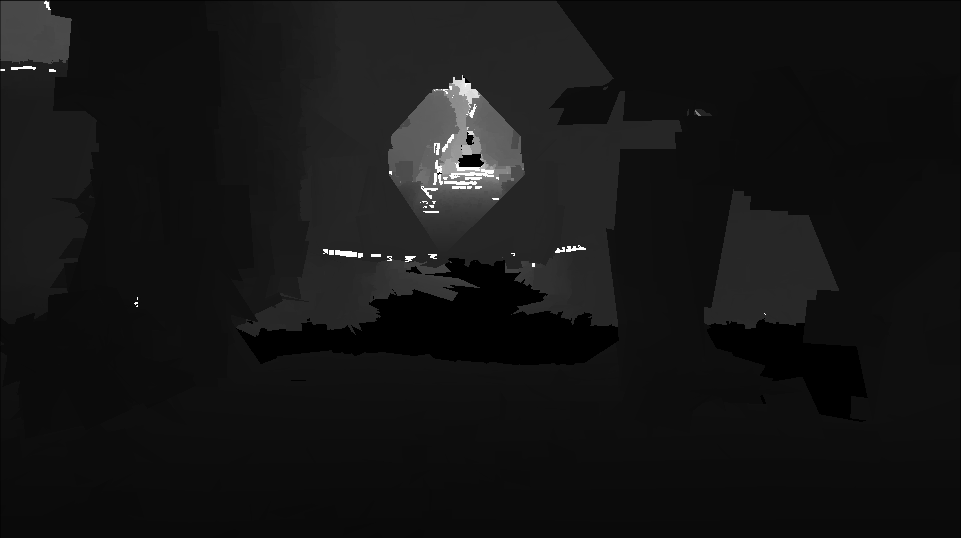}
    \end{tabular}
    \caption{Images and corresponding occlusion depth maps illustrating problems 
    coming
    from point cloud sparsity.}
    \label{fig:pcprob}
\end{figure}

As it can be seen in Figure~\ref{fig:manoir_map}, the point cloud seriously lacks completeness for parts that are not easily seen from an altitude of 30 meters. This includes vertical sections like walls or poles, like seen in Figure~\ref{fig:pcprob}, or cluttered sections, like under trees or 
%in 
inside
a tunnel. From this 
%statement, we can observe two things : 
remark, we can do two observations:
\begin{itemize}
    \item Although it covers a 
    %great 
    large
    area very quickly, UAV Lidar scanning does not cover a wide range of viewpoints. The 30 meters altitude makes it difficult to see the same thing as within a few meters of altitude. As such, it would have been useful to add Lidar scans from a lower altitude, from other methods. For example, we can use fixed Lidar and handheld Lidar. Regarding this use case, we recommend to use a handheld Lidar because it's much faster than fixed Lidar 
    %(it is as fast as a human walking 
    (typically as fast as human gait,
    while fixed Lidar requires several minutes per viewpoint), and if a UAV is available, all the hardware for using Lidar with IMU technique like \eg LIO-SAM \cite{liosam2020shan} or a proprietary technique like GeoSLAM is already available.
    \item In case no other Lidar scan is possible (\eg because the area is not reachable other than with a drone), the COLMAP cloud is locally good. An interesting compromise for future work would be to combine the Lidar cloud with COLMAP reconstruction for obstacles very close to the camera.
\end{itemize}

\subsubsection{Benchmark}
Scalar metrics are shown in Table \ref{tab:manoir_metrics}. 
Distributions of (estimation / ground truth) ratios (equivalent to 
log depth difference, top), 
%distributions and estimation-wise log error 
and log-errors according to the estimated depth (bottom)
are shown on Figure~\ref{fig:hists} (left panel), while samples can be seen Figure~\ref{fig:samples} (first three rows). This first evaluation shows that unsurprisingly, DepthNet largely outperforms SFMLearner in all metrics. This is also visible on the log difference distribution: although both distributions have their maximum at a null log difference (
%or 
i.e.
a ratio of 1), indicating an unbiased estimator, DepthNet's distribution is much more concentrated. We can also see that both distribution exhibits a clear skewness toward negative difference. This is corroborated with the second plot where we can see that for both algorithms, the log error is much lower above 20 meters. This is a good hint for a potential drone manufacturer to think of a dedicated system for low depth values. For example, a stereo camera setup would be complementary to these two systems, because it's more accurate for lower depth values.

For conciseness purpose, we have not shown all the histograms our tool can generate. However, we have made them easily generated within the tool we open sourced. \footnote{\url{https://github.com/ClementPinard/depth-dataset-builder\#depth-algorithm-evaluation}}
\begin{table*}
    \centering
\begin{tabular}{|l|c|c|c|c|c|c|c|c|c|c|}
\hline
Method & \cellcolor{blue!15}MAE & \cellcolor{blue!15}MRE & \cellcolor{blue!15}MLE & \cellcolor{blue!15}SAE & \cellcolor{blue!15}SLE & \cellcolor{red!25}$P_{1.25}$ & \cellcolor{red!25}$P_{1.25^2}$ & \cellcolor{red!25}$P_{1.25^3}$ \\
\hline
SFMLearner \cite{zhou2017unsupervised} & 18.40 & 0.5145 & 1.005 & 24.46 & 1.458 & 0.2395 & 0.4113 & 0.5385 \\
DepthNet \cite{pinard2018learning} & \textbf{11.99} & \textbf{0.3275} & \textbf{0.5290} & \textbf{18.51} & \textbf{0.9099} & \textbf{0.5241} & \textbf{0.7069} & \textbf{0.7717} \\
\hline
\end{tabular}
    \caption{Metric comparison 
    %of 
    between
    SFMLearner \cite{zhou2017unsupervised} and DepthNet \cite{pinard2018learning} on {\em Manoir} dataset.}
    \label{tab:manoir_metrics}
\end{table*}

\subsection{Second application: University hall dataset}
\subsubsection{Context}
\begin{figure}
    \centering
    \includegraphics[width=\columnwidth]{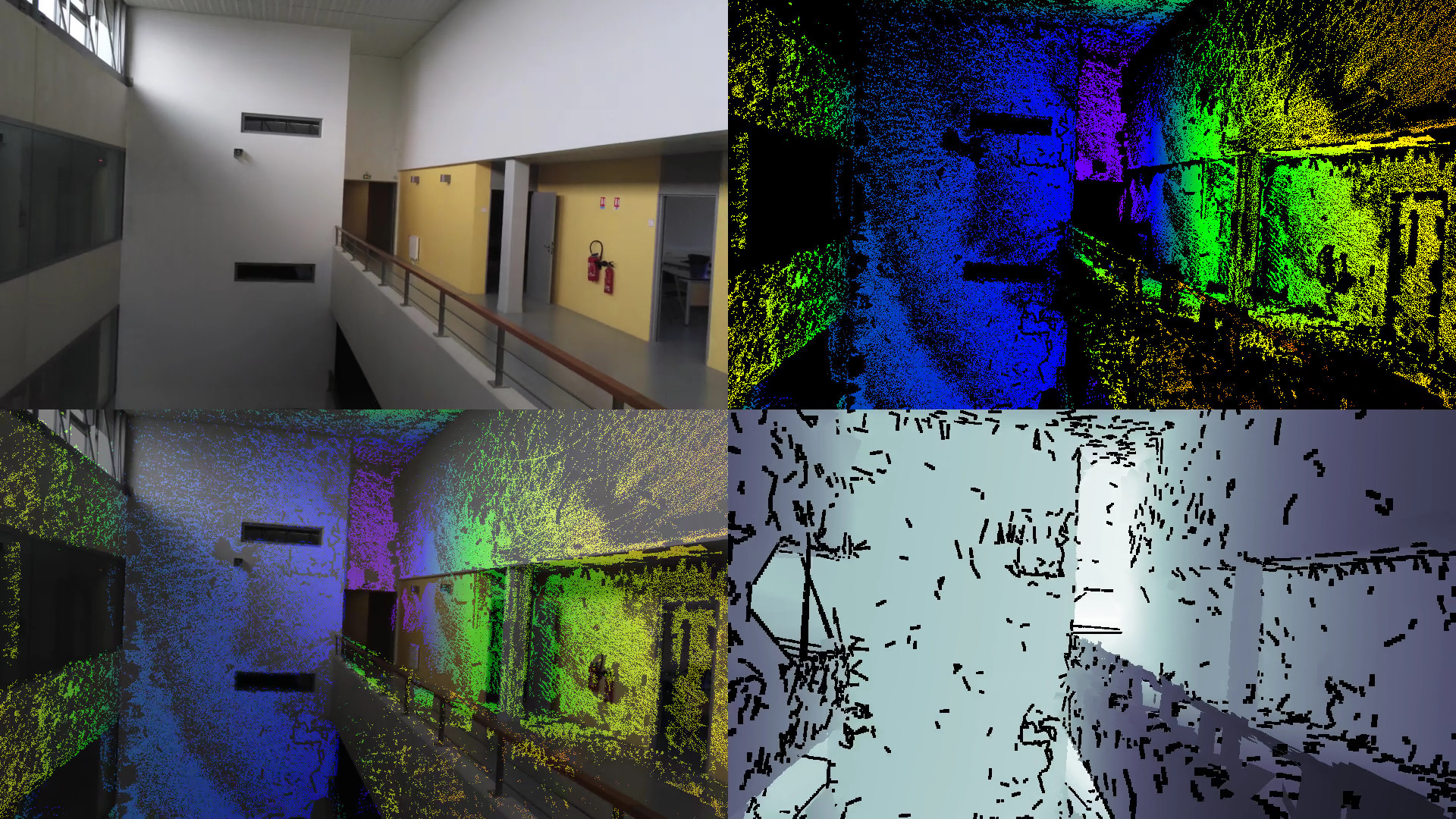}
    \caption{University hall depth visualization, same structure as Figure~\ref{fig:manoir_full}}
    \label{fig:ensta}
\end{figure}

\begin{figure}
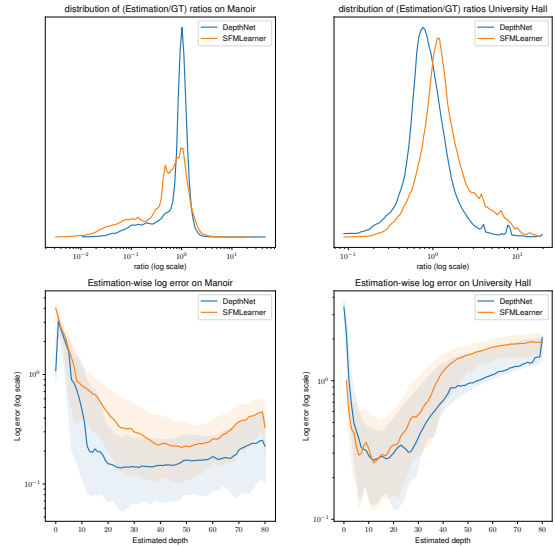

    \centering
    
    \resizebox{.45\columnwidth}{!}{
    \input{images/plots/my_graph_manoir.pgf}
    }
    \resizebox{.45\columnwidth}{!}{
    \input{images/plots/my_graph_ensta.pgf}
    }
    \caption{Comparison of error distributions between SFMLearner \cite{zhou2017unsupervised} and DepthNet \cite{pinard2018learning} on {\em Manoir} dataset (left) and on {\em University hall} dataset (right)
    Top:
    Distribution of ratios between estimation and ground truth (above 1 means estimation is higher), at log scale. This is equivalent to the log difference distribution.
    Bottom:
    %Estimation wise log error. 
    Log error as a function of estimated depth.
    For visualization purpose, the $y$ scale is logarithmic. The 
    %line 
    curve
    shows the median value and the shaded area represents the $ 50\% $ confidence interval.
    }
    \label{fig:hists}
\end{figure}

\begin{figure*}
    \centering
    \begin{tabular}{@{\hskip 0.1em}c@{\hskip 0.1em}c@{\hskip 0.1em}c@{\hskip 0.1em}c}
    Image & GroudTruth & DepthNet\cite{isprs-annals-IV-2-W3-67-2017} & SFMLearner \cite{zhou2017unsupervised} \\
    \includegraphics[width=.25\textwidth]{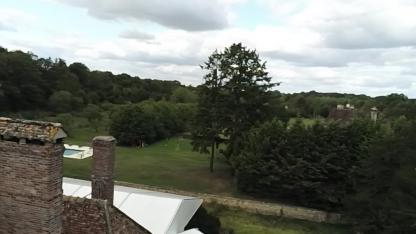} &
    \includegraphics[width=.25\textwidth]{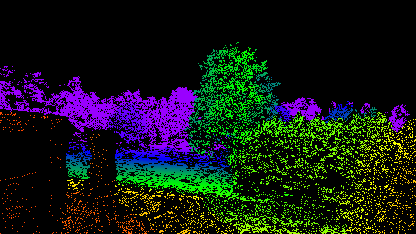} &
    \includegraphics[width=.25\textwidth]{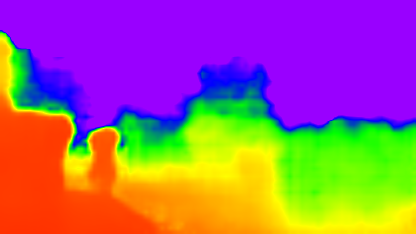} &
    \includegraphics[width=.25\textwidth]{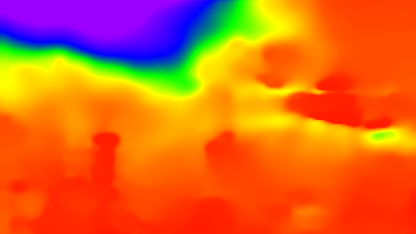} \\
    \includegraphics[width=.25\textwidth]{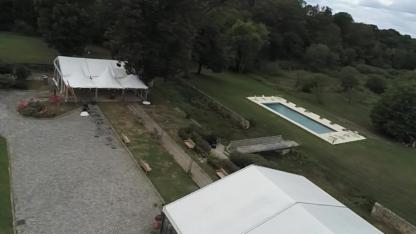} &
    \includegraphics[width=.25\textwidth]{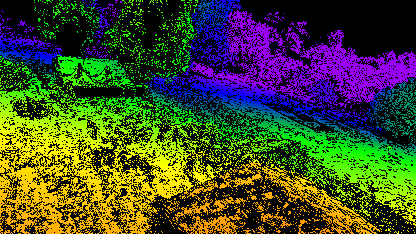} &
    \includegraphics[width=.25\textwidth]{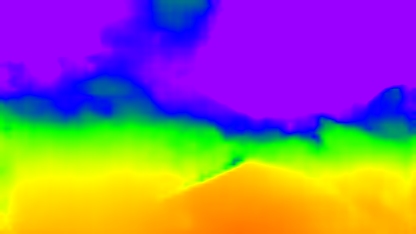} &
    \includegraphics[width=.25\textwidth]{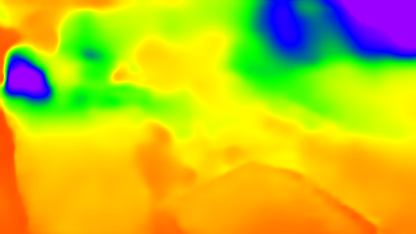} \\
    \includegraphics[width=.25\textwidth]{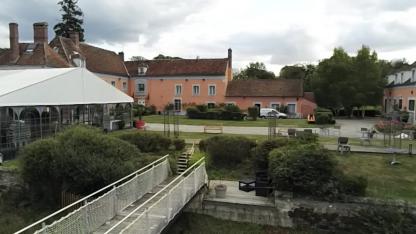} &
    \includegraphics[width=.25\textwidth]{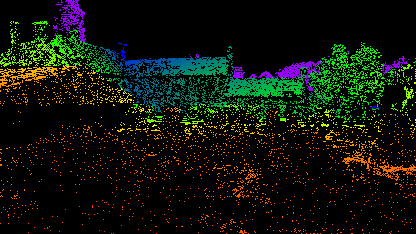} &
    \includegraphics[width=.25\textwidth]{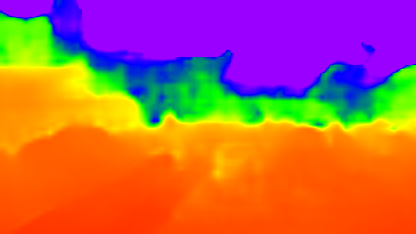} &
    \includegraphics[width=.25\textwidth]{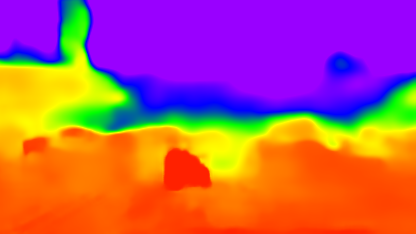} \\
    \includegraphics[width=.25\textwidth]{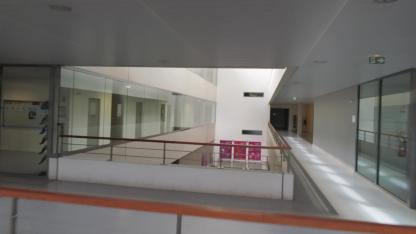} &
    \includegraphics[width=.25\textwidth]{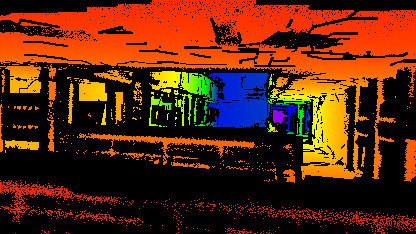} &
    \includegraphics[width=.25\textwidth]{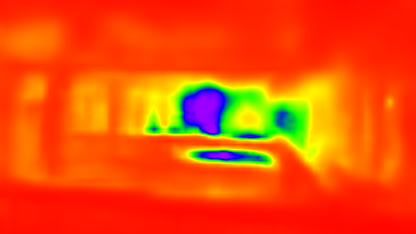} &
    \includegraphics[width=.25\textwidth]{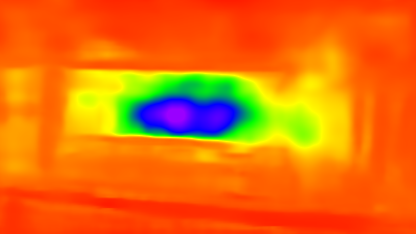} \\
    \includegraphics[width=.25\textwidth]{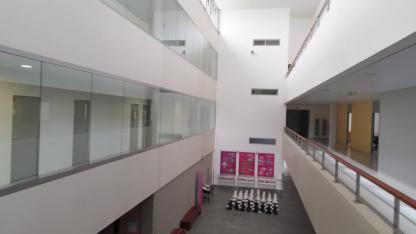} &
    \includegraphics[width=.25\textwidth]{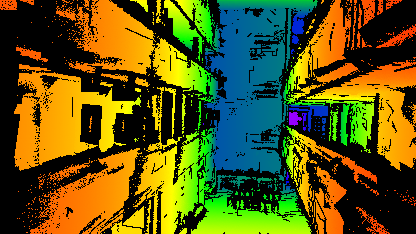} &
    \includegraphics[width=.25\textwidth]{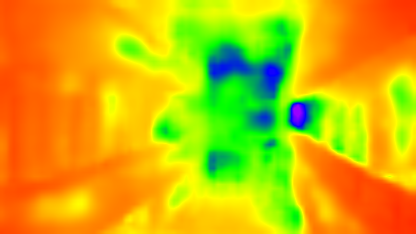} &
    \includegraphics[width=.25\textwidth]{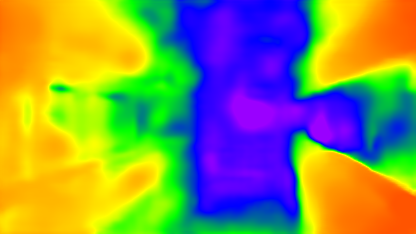}
    \end{tabular}
    \caption{Vizualisation of some samples on {\em Manoir} and {\em University hall} data sets. Colormap is 
    OpenCV
    Rainbow, normalized to Ground truth.}
    \label{fig:samples}
\end{figure*}

Our second 
%attempt 
use case
is an indoor scene, 
%presented 
shown on
Figure~\ref{fig:ensta}, the hall of a University with a very high ceiling. For scan mapping, we use a handheld Zeb Horizon from GeoSLAM, with an announced precision of 4 cm. For videos, we used an Anafi; we have no photogrammetry-oriented frames.
%As 
Like
for Manoir dataset, 
%see a 
the results of our 
small benchmark 
can be seen on
Table~\ref{tab:ensta_metrics} and Figure~\ref{fig:hists} (right panel). Some samples can be seen Figure~\ref{fig:samples} (last two rows)

Although DepthNet still outperforms SFMLearner for most metrics, the estimation accuracy is much lower. This can be explained by the fact that the university hall is mostly composed of completely white walls, with many glass windows and a somewhat reflective ground. This is a very challenging use case for SFM based training and inference methods. These results indicate that both SFMLearner and DepthNet are much more suited for outdoor scenarios, with irregular textured surfaces, and that 
%a structured light solution like Kinect is 
active sensing solutions like structured light or time-of-flight are
probably more interesting for this dataset.

\begin{table*}
    \centering
\begin{tabular}{|l|c|c|c|c|c|c|c|c|c|c|}
\hline
Method & \cellcolor{blue!15}MAE & \cellcolor{blue!15}MRE & \cellcolor{blue!15}MLE & \cellcolor{blue!15}SAE & \cellcolor{blue!15}SLE & \cellcolor{red!25}$P_{1.25}$ & \cellcolor{red!25}$P_{1.25^2}$ & \cellcolor{red!25}$P_{1.25^3}$ \\
\hline
SFMLearner \cite{zhou2017unsupervised} & 10.25 & 1.0225 & 0.5682 & 18.29 & 0.7798 & 0.3180 & 0.5461 & 0.6912\\
DepthNet \cite{pinard2018learning} & \textbf{5.890} & \textbf{0.5197} & \textbf{0.5056} & \textbf{10.612} & \textbf{0.726} & \textbf{0.3184} & \textbf{0.5982} & \textbf{0.7694} \\
\hline
\end{tabular}
    \caption{Metric comparison 
    %of 
    between
    SFMLearner \cite{zhou2017unsupervised} and DepthNet \cite{pinard2018learning} on {\em University hall} dataset}
    \label{tab:ensta_metrics}
\end{table*}

%AM2---> J'en suis là

\section{Discussion
and conclusive remarks}
In this paper, we presented 
RDC,
a powerful tool for depth dataset generation. It is very flexible 
with respect
to available budget and hardware. Its context of application can be as 
%lows as no budget at all, 
rudimentary as no-budget-at-all,
only using a handheld camera (\eg the user's phone camera), and 
still
it lets a user make the 
%best from 
most of
limited means, to have depth enabled videos with potentially infinite range. By 
%letting future users access it, 
leaving open access to future users,
we hope to improve it by 
%having a big picture of typical use cases.
gradually increasing the diversity of use cases. 

\subsection{Limitations}
%This tool unfortunately suffer from somes 
RDC still suffers from some
already known limitations:
\begin{itemize}
    \item The first limitation is obviously the 
    %necessity for a regid 
    need for a rigid
    scene. Although this problem can be mitigated in the case of a UAV flying, it can be particularly problematic for in-car environment in a urban area where many people and other vehicles are dynamic. It can also be a problem when Lidar scanning and video recording are not done at the same time, in a scene with rigid 
    but movable
    objects.
    %, but that can be moved. 
    This was the case in the University hall dataset, where some seats had been moved, thus 
    %rendering the
    making some
    depth maps useless. A possible solution would be to combine a dense high quality point cloud of a scene, scanned beforehand, and sparse point cloud from a mobile Lidar.  By characterizing possible dynamic elements, like humans and other vehicles in a urban area, we may be able to discard the depth map from rigid scene where the dynamic 
    %element is pictured, 
    elements are segmented,
    and replace it with the depth from the sparse mobile Lidar.
    \item Although it does not require heavy hardware for data acquisition, the computer used by this tool must be powerful (as 
    %it is the case for every 
    with any
    photogrammetry task), and have plenty of disk space, which mitigates the fact that we adapt to any kind of budget. However, a regular gaming desktop computer is often enough, and is orders of magnitude cheaper than a full Lidar and cam rig solution.
    \item Although it can be improved by using better registration algorithms than a simple ICP, the registration part will often need a human 
    %hand 
    eye
    to assess the registration quality. This part of the process lies in the middle of it, which makes the tool not totally automatic. It is essential that it happens after the photogrammetry and before ground truth generation, so there is no obvious solution yet.
    \item When the camera is located in a very sparse portion of the point cloud, the occlusion depth will be of poor quality. It is possible that the end user will discover after the whole dataset creation process that some scenes 
    %will 
    would
    need a better scanning process. It would 
    %certainly 
    then
    be interesting to 
    %have 
    enter
    a warning before the end of the process, as soon as the Lidar point cloud is registered, that some areas lack data.
\end{itemize}

\subsection{Further work}
The next logical step is now to construct a 
%full 
more comprehensive
dataset with this tool, using different scenes, contexts, and modalities, in the same way ETH3D dataset was 
%constructed, except 
built, but
with much more frames.

Additionally, a full evaluation suite would be very beneficial, as this 
%iteration 
initial version
only provides 
%very 
basic examples of histogram based algorithm quality assessment.

Finally, now that we have a way of localizing video frames in a point cloud, we can make 
%some 
more ground truths to test 
%possible algorithms to:
other kinds of algorithms:
\begin{itemize}
    \item From odometry and ground truth point cloud, ground truth 
    %of 
    optical flow can be deduced.
    \item If we annotate the Lidar point cloud with semantic parts, we can construct semantic segmentation maps.
\end{itemize}

\section*{Acknowledgements}
Acquisitions for the {\em Manoir} dataset were made in collaboration with AIRD'ECO-Drone\footnote{\url{https://www.airdeco-drone.com}} company, thanks to the financial support of Parrot\footnote{\url{https://www.parrot.com}} company.
Acquisitions for the {\em University hall} dataset were made entirely by Cl\'ement Pinard, thanks to the equipment and training provided by Geomesure\footnote{\url{https://www.geomesure.fr/}} company.

{\small
\bibliographystyle{ieee_fullname}
\bibliography{egbib}

\begin{thebibliography}{10}\itemsep=-1pt

\bibitem{121791}
P.~J. {Besl} and N.~D. {McKay}.
\newblock A method for registration of 3-d shapes.
\newblock {\em IEEE Transactions on Pattern Analysis and Machine Intelligence},
  14(2):239--256, 1992.

\bibitem{Butler:ECCV:2012}
D.~J. Butler, J. Wulff, G.~B. Stanley, and M.~J. Black.
\newblock A naturalistic open source movie for optical flow evaluation.
\newblock In {A. Fitzgibbon et al. (Eds.)}, editor, {\em European Conf. on
  Computer Vision (ECCV)}, Part IV, LNCS 7577, pages 611--625. Springer-Verlag,
  Oct. 2012.

\bibitem{CHEN1992145}
Yang Chen and Gerard Medioni.
\newblock Object modelling by registration of multiple range images.
\newblock {\em Image and Vision Computing}, 10(3):145 -- 155, 1992.
\newblock Range Image Understanding.

\bibitem{DFIB15}
A. Dosovitskiy, P. Fischer, E. Ilg, P. H{\"a}usser, C. Haz{\i}rba{\c{s}}, V.
  Golkov, P. v.d. Smagt, D. Cremers, and T. Brox.
\newblock Flownet: Learning optical flow with convolutional networks.
\newblock In {\em IEEE International Conference on Computer Vision (ICCV)},
  2015.

\bibitem{eigen2014depth}
David Eigen, Christian Puhrsch, and Rob Fergus.
\newblock Depth map prediction from a single image using a multi-scale deep
  network.
\newblock {\em Advances in neural information processing systems},
  27:2366--2374, 2014.

\bibitem{FuCVPR18-DORN}
Huan Fu, Mingming Gong, Chaohui Wang, Kayhan Batmanghelich, and Dacheng Tao.
\newblock {Deep Ordinal Regression Network for Monocular Depth Estimation}.
\newblock In {\em {IEEE Conference on Computer Vision and Pattern Recognition
  (CVPR)}}, 2018.

\bibitem{garg2016unsupervised}
Ravi Garg, BG~Vijay Kumar, Gustavo Carneiro, and Ian Reid.
\newblock Unsupervised cnn for single view depth estimation: Geometry to the
  rescue.
\newblock In {\em European Conference on Computer Vision}, pages 740--756.
  Springer, 2016.

\bibitem{geiger2012we}
Andreas Geiger, Philip Lenz, and Raquel Urtasun.
\newblock Are we ready for autonomous driving? the kitti vision benchmark
  suite.
\newblock In {\em 2012 IEEE Conference on Computer Vision and Pattern
  Recognition}, pages 3354--3361. IEEE, 2012.

\bibitem{monodepth2}
Cl{\'{e}}ment Godard, Oisin {Mac Aodha}, Michael Firman, and Gabriel~J.
  Brostow.
\newblock Digging into self-supervised monocular depth prediction.
\newblock In {\em The International Conference on Computer Vision (ICCV)},
  October 2019.

\bibitem{Gordon_2019_ICCV}
Ariel Gordon, Hanhan Li, Rico Jonschkowski, and Anelia Angelova.
\newblock Depth from videos in the wild: Unsupervised monocular depth learning
  from unknown cameras.
\newblock In {\em Proceedings of the IEEE/CVF International Conference on
  Computer Vision (ICCV)}, October 2019.

\bibitem{hartley_zisserman_2004}
Richard Hartley and Andrew Zisserman.
\newblock {\em Multiple View Geometry in Computer Vision}.
\newblock Cambridge University Press, 2 edition, 2004.

\bibitem{kazhdan2006poisson}
Michael Kazhdan, Matthew Bolitho, and Hugues Hoppe.
\newblock Poisson surface reconstruction.
\newblock In {\em Proceedings of the fourth Eurographics symposium on Geometry
  processing}, volume~7, 2006.

\bibitem{Knapitsch2017}
Arno Knapitsch, Jaesik Park, Qian-Yi Zhou, and Vladlen Koltun.
\newblock Tanks and temples: Benchmarking large-scale scene reconstruction.
\newblock {\em ACM Transactions on Graphics}, 36(4), 2017.

\bibitem{Photogrammetry}
Karl Kraus, Ian~A. Harley, and Stephen Kyle.
\newblock {\em Photogrammetry: Geometry from Images and Laser Scans}.
\newblock De Gruyter, Berlin, Boston, 18 Oct. 2011.

\bibitem{VOT_TPAMI}
Matej Kristan, Jiri Matas, Ale\v{s} Leonardis, Tomas Vojir, Roman Pflugfelder,
  Gustavo Fernandez, Georg Nebehay, Fatih Porikli, and Luka \v{C}ehovin.
\newblock A novel performance evaluation methodology for single-target
  trackers.
\newblock {\em IEEE Transactions on Pattern Analysis and Machine Intelligence},
  38(11):2137--2155, Nov 2016.

\bibitem{delauney}
P. Labatut, J.-P. Pons, and R. Keriven.
\newblock {Robust and Efficient Surface Reconstruction From Range Data}.
\newblock {\em Computer Graphics Forum}, 2009.

\bibitem{lee2019big}
Jin~Han Lee, Myung-Kyu Han, Dong~Wook Ko, and Il~Hong Suh.
\newblock From big to small: Multi-scale local planar guidance for monocular
  depth estimation.
\newblock {\em arXiv preprint arXiv:1907.10326}, 2019.

\bibitem{lopez2017aggressive}
Brett~T Lopez and Jonathan~P How.
\newblock Aggressive 3-d collision avoidance for high-speed navigation.
\newblock In {\em 2017 IEEE International Conference on Robotics and Automation
  (ICRA)}, pages 5759--5765. IEEE, 2017.

\bibitem{Macqueen67somemethods}
J. Macqueen.
\newblock Some methods for classification and analysis of multivariate
  observations.
\newblock In {\em In 5-th Berkeley Symposium on Mathematical Statistics and
  Probability}, pages 281--297, 1967.

\bibitem{nalpantidis2009stereovision}
Lazaros Nalpantidis, Ioannis Kostavelis, and Antonios Gasteratos.
\newblock Stereovision-based algorithm for obstacle avoidance.
\newblock In {\em International Conference on Intelligent Robotics and
  Applications}, pages 195--204. Springer, 2009.

\bibitem{Silberman:ECCV12}
Pushmeet~Kohli Nathan~Silberman, Derek~Hoiem and Rob Fergus.
\newblock Indoor segmentation and support inference from rgbd images.
\newblock In {\em ECCV}, 2012.

\bibitem{pinardphd}
Cl{\'e}ment Pinard.
\newblock {\em {Robust Learning of a depth map for obstacle avoidance with a
  monocular stabilized flying camera}}.
\newblock Theses, {Universit{\'e} Paris Saclay (COmUE)}, June 2019.

\bibitem{isprs-annals-IV-2-W3-67-2017}
C. Pinard, L. Chevalley, A. Manzanera, and D. Filliat.
\newblock End-to-end depth from motion with stabilized monocular videos.
\newblock {\em ISPRS Annals of Photogrammetry, Remote Sensing and Spatial
  Information Sciences}, IV-2/W3:67--74, 2017.

\bibitem{pinard2018learning}
Cl{\'e}ment Pinard, Laure Chevalley, Antoine Manzanera, and David Filliat.
\newblock Learning structure-from-motion from motion.
\newblock In {\em Proceedings of the European Conference on Computer Vision
  (ECCV)}, 2018.

\bibitem{doi:10.1021/ac60214a047}
Abraham. Savitzky and M.~J.~E. Golay.
\newblock Smoothing and differentiation of data by simplified least squares
  procedures.
\newblock {\em Analytical Chemistry}, 36(8):1627--1639, 1964.

\bibitem{saxena20083}
Ashutosh Saxena, Sung~H Chung, and Andrew~Y Ng.
\newblock 3-d depth reconstruction from a single still image.
\newblock {\em International journal of computer vision}, 76(1):53--69, 2008.

\bibitem{Schilling_MG-BDDPBL}
Hendrik Schilling, Marcel Gutsche, Alexander Brock, Dane Sp{\"a}th, Carsten
  Rother, and Karsten Krispin.
\newblock Mind the gap - a benchmark for dense depth prediction beyond lidar.
\newblock In {\em 2020 IEEE Conference on Computer Vision and Pattern
  Recognition Workshops (CVPRW)}, volume in press, 2020.

\bibitem{schoenberger2016sfm}
Johannes~Lutz Sch\"{o}nberger and Jan-Michael Frahm.
\newblock Structure-from-motion revisited.
\newblock In {\em Conference on Computer Vision and Pattern Recognition
  (CVPR)}, 2016.

\bibitem{schoenberger2016vote}
Johannes~Lutz Sch\"{o}nberger, True Price, Torsten Sattler, Jan-Michael Frahm,
  and Marc Pollefeys.
\newblock A vote-and-verify strategy for fast spatial verification in image
  retrieval.
\newblock In {\em Asian Conference on Computer Vision (ACCV)}, 2016.

\bibitem{schoenberger2016mvs}
Johannes~Lutz Sch\"{o}nberger, Enliang Zheng, Marc Pollefeys, and Jan-Michael
  Frahm.
\newblock Pixelwise view selection for unstructured multi-view stereo.
\newblock In {\em European Conference on Computer Vision (ECCV)}, 2016.

\bibitem{schoeps2017cvpr}
Thomas Sch\"ops, Johannes~L. Sch\"onberger, Silvano Galliani, Torsten Sattler,
  Konrad Schindler, Marc Pollefeys, and Andreas Geiger.
\newblock A multi-view stereo benchmark with high-resolution images and
  multi-camera videos.
\newblock In {\em Conference on Computer Vision and Pattern Recognition
  (CVPR)}, 2017.

\bibitem{liosam2020shan}
Tixiao Shan, Brendan Englot, Drew Meyers, Wei Wang, Carlo Ratti, and Rus
  Daniela.
\newblock Lio-sam: Tightly-coupled lidar inertial odometry via smoothing and
  mapping.
\newblock In {\em IEEE/RSJ International Conference on Intelligent Robots and
  Systems (IROS)}, pages 5135--5142. IEEE, 2020.

\bibitem{6361384}
G.~K.~L. {Tam}, Z. {Cheng}, Y. {Lai}, F.~C. {Langbein}, Y. {Liu}, D.
  {Marshall}, R.~R. {Martin}, X. {Sun}, and P.~L. {Rosin}.
\newblock Registration of 3d point clouds and meshes: A survey from rigid to
  nonrigid.
\newblock {\em IEEE Transactions on Visualization and Computer Graphics},
  19(7):1199--1217, 2013.

\bibitem{Uhrig2017THREEDV}
Jonas Uhrig, Nick Schneider, Lukas Schneider, Uwe Franke, Thomas Brox, and
  Andreas Geiger.
\newblock Sparsity invariant cnns.
\newblock In {\em International Conference on 3D Vision (3DV)}, 2017.

\bibitem{diode_dataset}
Igor Vasiljevic, Nick Kolkin, Shanyi Zhang, Ruotian Luo, Haochen Wang,
  Falcon~Z. Dai, Andrea~F. Daniele, Mohammadreza Mostajabi, Steven Basart,
  Matthew~R. Walter, and Gregory Shakhnarovich.
\newblock {DIODE}: {A} {D}ense {I}ndoor and {O}utdoor {DE}pth {D}ataset.
\newblock {\em CoRR}, abs/1908.00463, 2019.

\bibitem{zhou2017unsupervised}
Tinghui Zhou, Matthew Brown, Noah Snavely, and David~G. Lowe.
\newblock Unsupervised learning of depth and ego-motion from video.
\newblock In {\em CVPR}, 2017.

\end{thebibliography}
}
\end{document}